\documentclass[twoside]{article}

\usepackage[accepted]{aistats2022}
\usepackage[round]{natbib}
\renewcommand{\refname}{References}

\newcommand{\diag}{\text{diag}}

\usepackage{graphicx}
\usepackage[utf8]{inputenc}
\usepackage[british]{babel}
\usepackage{xcolor}
\usepackage{amssymb}
\usepackage{bm}
\usepackage{amsmath}
\usepackage{mathtools}

\usepackage[bookmarksnumbered=true, bookmarksopen=true, bookmarksopenlevel=1, colorlinks=false, pdfstartview=Fit, pdfpagemode=UseOutlines]{hyperref}
\AtBeginDocument{}
\usepackage{hypcap}

\usepackage{algorithm}
\usepackage{algpseudocode}

\newtheorem{theorem}{Theorem}%[section]
\newtheorem{corollary}{Corollary}[theorem]
\newtheorem{lemma}[theorem]{Lemma}

\usepackage{caption}

\begin{document}



\twocolumn[

\aistatstitle{Safe Active Learning for Multi-Output Gaussian Processes}

\aistatsauthor{ Cen-You Li \And Barbara Rakitsch \And Christoph Zimmer }

\aistatsaddress{Cen-You.Li@de.bosch.com \And Barbara.Rakitsch@de.bosch.com \\ Bosch Center for Artificial Intelligence \\ Robert-Bosch-Campus 1, 71272 Renningen, Germany \And Christoph.Zimmer@de.bosch.com} ]

\begin{abstract}
	Multi-output regression problems are commonly encountered in science and engineering. 
	In particular, multi-output Gaussian processes have been emerged as a promising tool for modeling these complex systems since they can exploit the inherent correlations and provide reliable uncertainty estimates. 
	In many applications, however, acquiring the data is expensive and safety concerns might arise (e.g. robotics, engineering). 
	We propose a safe active learning approach for multi-output Gaussian process regression. 
	This approach queries the most informative data or output taking the relatedness between the regressors and safety constraints into account. 
	We prove the effectiveness of our approach by providing theoretical analysis and by demonstrating empirical results on simulated datasets and on a real-world engineering dataset. 
	On all datasets, our approach shows improved convergence compared to its competitors.
\end{abstract}

\section{Introduction}\label{section-intro}

Active learning (AL) selects the most informative data sequentially according to previous measurements and an acquisition function~\citep{krause08a, journals/corr/abs-1112-5745, AAAI1611879}. 
The objective is to optimize a model without labeling unnecessary data. 
The problem setup is closely related to Bayesian optimization, i.e. BO~\citep{brochu2010tutorial}, which optimizes a black-box function with limited exploration. 
In various scenarios, safety concerns are also critical during the exploration phase. 
For instance, movements of a machine are not supposed to crash any objects. 
A system should avoid generating high pressure, high temperature, or explosion. 
Safe learning addresses this by incorporating and learning safety constraints~\citep{pmlr-v37-sui15}.
~\cite{10.1007/978-3-319-23461-8_9} and~\cite{NEURIPS2018_b197ffde} combine safety considerations with AL so that the data selection is done only in the determined safe domain.

These works, however, rarely considered multi-output (MO) regression problems, despite them commonly encountered in science, engineering and medicine~\citep{xu2019survey, zhang2021survey, LIU2018102}. 
In such problems, it is possible to consider individual tasks or outputs independently, but the plausibly shared mechanisms are ignored, and the performances or data efficiency might be deteriorated.
~\cite{AAAI1611879} dealt with AL on MO models but focused on efficient computation of AL with large datasets and safe exploration was not addressed. 

We consider safe AL for MO regression models that exploit the correlations. 
In particular, we focus on problems in which different output components may not be synchronously observed (e.g. due to different measuring cost or difficulty). 
MO Gaussian processes (GPs) are natural candidates for these problems~\citep{NIPS2007_66368270, JMLR:v12:alvarez11a, alvarez2012kernels, vanderwilk2020framework}, due to their capability of capturing the correlations among different outputs and of quantifying the uncertainty. 

In our work, we consider as main model the Linear Model of Coregionalization (LMC,~\cite{etde5214736}), in which each output is modeled as a weighted sum of shared latent functions. Each latent function is drawn from a GP.
Later on, we extend the theoretical analysis also to the convolution process~\citep{higdon2002space, JMLR:v12:alvarez11a} in which each latent function is additionally convolved by an output-specific smoothing kernel.

To the best of our knowledge, this is the first framework about safe AL for MOGP regression. Our contributions can be summarized as follows:
\begin{itemize}
	\item We formulate an acquisition function for safe active learning in the MOGP framework that allows asynchronous measurements.
	\item We provide theoretical analysis of the safe AL algorithm in our framework, particularly we derive a convergence rate to the algorithm.
	\item We demonstrate the performance and superiority to state-of-the-art competitors on a real-world engineering dataset.
\end{itemize}

The overview of this papers is as follows. In section~\ref{section-related_work}, we briefly review the related works. 
In section~\ref{section-methods}, we introduce our algorithm.
We discuss the theory behind our algorithm in section~\ref{section-thms}, and validate empirically its usefulness in section~\ref{section-experiments}.
Finally, section~\ref{section-conclusion} concludes our work. 

\section{Related Work}\label{section-related_work}

AL has been extensively investigated for classification tasks~\citep{Hoi_ICML2006, 5206627, journals/corr/abs-1112-5745, GCAI2019:On_Robustness_of_Active, pmlr-v130-shi21a}, but less literature addresses AL in the regression setting~\citep{krause08a, GarnettOH2013}.
The problem setup is closely related to BO~\citep{brochu2010tutorial}. 
While AL and BO both consider limited exploration, the goals are very different.
AL aims to obtain a well performing model, usually with characteristic of overall precision, but BO only finds an optimum, e.g. a configuration of best performance or lowest cost.
In a BO problem, the model quality for points far away from the optimum is not important and can be really bad.
For a more general problem in this line of research, i.e. optimizing under uncertainty, GPs, which are capable of making predictions under uncertainty, are often used as surrogate models~\citep{brochu2010tutorial, Srinivas_2012}.

In recent years, the importance of safety considerations has led to a novel line of research ranging from Safe Bayesian optimization~\citep{pmlr-v37-sui15, Berkenkamp_2016, berkenkamp2020bayesian} to safe AL in a static environment~\citep{10.1007/978-3-319-23461-8_9} or dynamic systems~\citep{NEURIPS2018_b197ffde}. 
None of these contributions, however, consider MO which is able to exploit correlations among outputs.

Exploiting MO correlations has been shown successful in various applications~\citep{casale2017joint, LIU2018102, s12911-020-1069-4}. 
State of the art MO models, in particular with GPs (also refer to~\cite{alvarez2012kernels} and~\cite{vanderwilk2020framework} for an overview over MOGPs), include the Linear Model of Coregionalization (LMC), a simple yet effective model~\citep{etde5214736, NIPS2007_66368270, teh2005semiparametric}, and one of its extensions, the convolution process, which further captures correlations in multiple outputs that vary in smoothness~\citep{higdon2002space, JMLR:v12:alvarez11a}.

Complexity of GPs and MOGPs scales cubically with the number of observations~\citep{3569}. 
Existing works focus a lot on approximation methods for large datasets with GPs~\citep{pmlr-v5-titsias09a, hensman2013gaussian} and with MOGPs~\citep{JMLR:v12:alvarez11a, Nguyen2014UAI_CMOGP, vanderwilk2020framework}. 
In contrast to~\cite{AAAI1611879}, both our simulation and our real-world dataset can be modeled with very few data points. 
We thus consider MOGPs without any sparse approximations, even though these methods could be incorporated into our approach.

Very few works have tried to combine MO modeling with BO~\citep{NIPS2013_f33ba15e} or AL~\citep{AAAI1611879}.
~\cite{NIPS2013_f33ba15e} focused on transferring BO results between tasks, while~\cite{AAAI1611879} investigated efficient computation of AL for sparse MOGPs ~\citep{AAAI1611879}. 
To the best of our knowledge, none of the literature addressed safe data query or safe AL for MOGPs. 

\section{Methods}\label{section-methods}

We first provide background on GPs and MOGPs, and different inference strategies. 
In a second step, we show how safe active learning can be applied over multiple outputs.

\subsection{GP Regression}\label{section-GP}
\paragraph{Single-output}

A GP is a stochastic process where every finite subset follows a multivariate normal distribution. 
In GP regression, given observed data $\mathcal{D}=\{\bm{x}_{n} \in \mathbb{R}^D, y_{n} \in \mathbb{R}\}_{n=1}^{N}$, we specify a mean function $m: \mathbb{R}^D \rightarrow \mathbb{R}$ and a positive definite kernel function (covariance function) $k: \mathbb{R}^D \times \mathbb{R}^D \rightarrow \mathbb{R}$ as a GP prior for the function. 
The observations $y_n$ are assumed to be the functional values blurred by
%independent and identically distributed
i.i.d. Gaussian noise. 
The model is formulated as
\begin{equation*}g \sim \mathcal{GP} \left( m(\cdot), k(\cdot, \cdot) \right), y_{n}=g(\bm{x}_{n}) + \epsilon_{n}, \epsilon_{n} \sim \mathcal{N} \left( 0, \sigma^2 \right) .\end{equation*}
The goal is to predict $g(\bm{x}_*)$ and its uncertainty for a new input $\bm{x}_{*}$.
Assuming for simplicity a zero mean prior, $m \equiv 0$, the posterior is $p \left( g(\bm{x}_{*}) | \bm{x}_{*}, \mathcal{D} \right) = \mathcal{N} \left( \mu(\bm{x}_{*}), var(\bm{x}_{*}) \right) $, with
\begin{align}\label{eqn-gp_mean}
\mu(\bm{x}_{*}) &= K_{N*}^T \left( K_{NN} + \sigma^2 I \right)^{-1} (y_1, ..., y_N)^T,
\\
\label{eqn-gp_variance}
var(\bm{x}_{*}) &= k(\bm{x}_{*}, \bm{x}_{*}) - K_{N*}^T \left( K_{NN} + \sigma^2 I \right) ^{-1} K_{N*},
\end{align}
where $K_{N*} \in \mathbb{R}^{N\times1}$ and $K_{NN} \in \mathbb{R}^{N \times N}$ are matrices with $[K_{N*}]_{i} = k(\bm{x}_{i}, \bm{x}_{*})$ and $[K_{NN}]_{i, j} = k(\bm{x}_{i}, \bm{x}_{j})$. 
For further details, please see~\cite{3569}.

\paragraph{Multi-output (MO)}
We consider LMC as our main model~\citep{etde5214736}. 
Here we have $\bm{y}_n=\bm{f}(\bm{x}_n)+\bm{\epsilon}_n = W\bm{g}(\bm{x}_n)+\bm{\epsilon}_n \in \mathbb{R}^{P}$ with i.i.d. noise $[\bm{\epsilon}_n]_{p} \sim \mathcal{N} \left( 0, \sigma_{p}^{2} \right) $ for $p=1, 2, ..., P$, linear transformation $W \in \mathbb{R}^{P \times L}$, and latent GPs $g_l(\cdot) = [\bm{g}(\cdot)]_{l} \sim \mathcal{GP} \left( 0, k_l(\cdot, \cdot) \right) $ for $l=1, ..., L$.

Throughout this paper, we further assume finite $P$, finite $L$, bounded $k_l(\cdot, \cdot)$, and each element of $W$ bounded by a constant. Let $f_p(\cdot) = [\bm{f}(\cdot)]_{p}$. 
In this model, $\{f_{p}(\bm{x})\}_{p=1}^{P}$ is also a GP where every finite subset has zero mean and covariance
$cov \left( f_{p}(\bm{x}), f_{p'}(\bm{x}') \right)=$ 
$%\begin{equation}\label{def-mo_covariance}
\sum_{l=1}^{L} W_{pl}W_{p'l} k_{l}(\bm{x}, \bm{x}') \eqqcolon \eta_{p, p'}(\bm{x}, \bm{x}').
$%\end{equation}

Let $\bm{Y}$ denote the collection of observations $\{\bm{y}_{n} \in \mathbb{R}^{P}\}_{n=1}^{N}$, $\bm{X}$ denote $\{\bm{x}_{n}\}_{n=1}^{N}$, and $\mathcal{D} = \{ \bm{X}, \bm{Y} \}$. 
Let $y_{pn}$ be the $p$-th component of the $n$-th observation, i.e. $y_{pn} = [\bm{y}_{n}]_{p}$. 
The posterior $p(\bm{f}(\bm{x}_{*}) | \bm{x}_{*}, \mathcal{D})$ is a multivariate Gaussian $\mathcal{N}(\mu(\bm{x}_{*}), \Sigma(\bm{x}_{*}))$ with
\begin{align}
\label{eqn-mo_posterior_mu}
\mu(\bm{x}_{*}) &= \Omega_{N*}^T \left( \Omega_{NN} + \diag(\{\sigma_i^2\}_{p=1}^{P}) \otimes I_N \right) ^{-1} \bm{Y},
\\
\label{eqn-mo_posterior_cov} 
\Sigma(\bm{x}_{*}) &= \Omega_{**} - \Omega_{N*}^T \left( \Omega_{NN} + \diag(\{\sigma_i^2\}_{p=1}^{P}) \otimes I_N \right) ^{-1} \Omega_{N*},
\end{align}
where $\otimes$ denotes the Kronecker product, $\Omega_{**}$, $\Omega_{N*}$ and $\Omega_{NN}$ are gram matrices of kernel $\eta_{p, p'}(\cdot, \cdot)$. 
See our supplementary section~\ref{supp_section-MOGP} for full expression of the matrices and for the derivation of this posterior.

%Here, the complexity is dominated by the inverse which takes $\mathcal{O}(P^3N^3)$ time. 
Notice that $\bm{Y}$ can be ordered differently, but the corresponding permutation needs to be applied to the current $\Omega_{N*}$, $\Omega_{NN}$ and $\diag(\{\sigma_i^2\}_{p=1}^{P}) \otimes I_N$. 
As the permutation matrices cancel each other out, the posterior stays in the same form with only different indexing.

\paragraph{Partially observed MO}
In the previous section, each observation of $\bm{Y}$ has every component observed  for every input $\bm{X}$. 
In the following, we assume that some components can be omitted to save measuring costs as well as computational costs due to smaller $\Omega_{\cdot \cdot}$. 
Let $N_p$ be the number of outputs with $p$-th component observed and $N_{sum} = \sum_{p=1}^{P} N_p$. 
If the output is fully observed, we can see that $N_1=...=N_P=N$ and $N_{sum} = PN$.

For clarification, we define a reindexing bijection that maps the original index pairs to scalar indices (i.e. the scheme concatenates the outputs over all components into an one-dimensional vector), and the non-observed components are assigned with negative or zero indices. 
With this bijection, we can consider all observed output components by looking only at the positive indices ranging from $1$ to $N_{sum}$.

The notation of outputs now becomes $\bm{Y}_{\phi} = \{y_{p_{k}n_{k}}\}_{k=1}^{N_{sum}}$, where $\phi:(p, n) \rightarrow k$ is a re-indexing bijection with $(p_{k}, n_{k}) = \phi^{-1}(k)$. 
The output domain of $\phi$ is 
$\mathbb{Z} \cap [-NP+N_{sum}+1, N_{sum}]$
%$\{k \in \mathbb{Z} | -NP+N_{sum}+1 \leq k \leq N_{sum}\}$
, where 
$\{1, ..., N_{sum}\}$
%$\{k \in \mathbb{Z} | 1 \leq k \leq N_{sum}\}$
are the new indices of all observed output components. 
Notice that the notation is adapted from the fully observed scenario, so $\phi$ is dependent of $N$ (i.e. we clearly have $N_p \leq N, \forall p$). However, we omit $N$ in the notation for simplicity.

In addition, the corresponding rows and columns of the gram matrices are also omitted, and when we make predictions for one output component, the notation becomes as follows:
\begin{align}\label{eqn-mo_posterior_mu_poo} \mu(\bm{x}_{*}, p_*) &=
[\Omega_{N_{sum}*}]_{all,p*}^T
\widehat{\Omega}_{N_{sum}N_{sum}}^{-1}
\bm{Y}_{\phi},
\\
\label{eqn-mo_posterior_cov_poo}
\begin{split}
\Sigma(\bm{x}_{*}, p_*) &= \eta_{p_*, p_*}(\bm{x}_{*}, \bm{x}_{*})\\
&- [\Omega_{N_{sum}*}]_{all,p*}^T \widehat{\Omega}_{N_{sum}N_{sum}}^{-1} [\Omega_{N_{sum}*}]_{all,p*},
\end{split}
\end{align}
where $\widehat{\Omega}_{N_{sum}N_{sum}} = \Omega_{N_{sum}N_{sum}} + diag(\{\sigma_{p_k}^2\}_{k=1}^{N_{sum}})$.
Further notice that we omit the components without changing the order, so $\{\sigma_{p_k}^2\}_{k=1}^{N_{sum}}$ is actually $\{\sigma_1^2\}_{i=1}^{N_1}$ followed by $\{\sigma_2^2\}_{i=1}^{N_2}$ and so on.

\paragraph{MOGPs v.s. Multiple independent single-output GPs}
Concatenating $P$ single output GPs without modeling the output correlations is equivalent to a MOGP with $W = I_p$, i.e. $\eta_{p, p} = k_p$ and $\eta_{p, p'} \equiv 0$ for $p \neq p'$. 
Notice that in this case, the gram matrix $\Omega_{N_{sum},N_{sum}}$ and the inverse with noise variances have only non-zero components on the diagonal subblocks corresponding to $\eta_{p, p}$. 
The cross-covariance $\Omega_{N_{sum},*}$ has only non-zero components at $(iN+p, p)$ entries, $i=0,...,P-1$, and thus the posterior is identical to squeezing the posteriors of individual GPs into one vector/matrix. See supplementary section~\ref{supp_section-MOGP} for full matrix expression. 

On the other hand, if $W \neq I_P$, information can flow between the components which can ultimately lead to more accurate predictions and smaller uncertainty estimates.
%On the other hand, if $W \neq I_P$, even though the outputs are only partially observed, eq.~\eqref{eqn-mo_posterior_mu_poo}-\eqref{eqn-mo_posterior_cov_poo} show that individual component of observations contributes to posterior predictions of other output components.

\subsection{Inference with Hyperparameters}

The choice of kernel(s) and noise variance(s) allows the model to express various patterns learned from the data. 
This, however, requires the tuning of hyperparameters, jointly denoted by $\bm{\theta}$.

\paragraph{Type II maximum likelihood estimation}

A simple way is to select hyperparameters that maximize the log marginal likelihood. Mathematical detail is provided in supplementary section~\ref{supp_section-exp_detail_hyperparameters_tuning_detail}.
%Computing the likelihood (eq.~\ref{supp_obj-mo_log_marginal_likelihood}) has complexity $\mathcal{O}(N_{sum}^3)$ as the inversion of the covariance matrices is required.
Note that GPs are not scalable to large datasets without any approximations such as sparse variational inference~\citep{pmlr-v5-titsias09a, hensman2013gaussian}.
Such kind of approximation techniques could also be incorporated into our MOGP model~\citep{Nguyen2014UAI_CMOGP, vanderwilk2020framework}. 

\paragraph{Bayesian treatment}
Maximum likelihood estimation can suffer from overfitting problems.
This in particular holds true for the low-data regime in which we are operating.
On the contrary, we can assign prior distributions over the hyperparameters and compute the predictive GP posterior over all possible hyperparameters.
This inference is then an integral over the hyperparameters, which is intractable.
We either need to perform approximate inference~\citep{titsias2014doubly} or resort to Monte Carlo sampling. 
In our work, we apply the latter.
We use Hamiltonian Monte Carlo (HMC)~\citep{betancourt2018conceptual, brooks2011handbook} as our sampling method.
Extensions to sparse GPs also exist~\citep{hensman2015mcmc}.
We refer to section~\ref{supp_section-exp_detail_bayesian_treatment} for mathematical detail.

\subsection{Safe AL}\label{section-safety_AL}

Our algorithm extends the work of~\cite{NEURIPS2018_b197ffde} to the multi-output scenario.
The general goal of AL is to obtain good models with as few data points as possible.
AL methods are especially important when it is expensive to measure training data (e.g. expensive to hire an expert or run large devices).
The model performance can be quantified e.g. by uncertainty or by RMSE.
Here we introduce the algorithm we use, and then in 
section~\ref{section-thms} we show that the uncertainty of the model decreases to zero with the safe AL.

\paragraph{Pool-based AL}
AL is a sequential learning scheme that allows us to query only the most informative data for a problem.
In each learning iteration, we are given an observed dataset $\mathcal{D}$ and a pool set $\mathcal{D}_{pool}$ containing candidate points that can be queried.
We query a new observation $\bm{y}_a(\bm{x}_a)$ from the pool according to an acquisition function $\alpha, \alpha(\cdot) \in \mathbb{R}$ such that 
%\begin{equation}\label{prob-general_al}
$\bm{x}_a = \text{argmax}_{\bm{x}} \{\alpha(\bm{x}, \mathcal{D}) | \bm{x} \in \mathcal{D}_{pool}\}$.
%\end{equation}
The acquisition function determines the gain of acquiring each candidate without access to the actual $\bm{y}$ value corresponding to this candidate.
In a real application, data that are not queried would not be measured.
After the query, the corresponding new measurement $\bm{y}_a$ will be provided.
Therefore, the observed and pool sets become $\mathcal{D} \cup \{\bm{x}_a, \bm{y}_a\}$ and $\mathcal{D}_{pool} \setminus \{\bm{x}_a, \bm{y}_a\}$ respectively, and the new iteration is conducted with the updated datasets. 
When the outputs are partially observed, $\mathcal{D}_{pool} = \{((\bm{x}, p), y_p)\}$ and the query problem is $(\bm{x}_a, p_a) = \text{argmax}_{\bm{x}, p} \{\alpha(\bm{x}, p, \mathcal{D}) | (\bm{x}, p) \in \mathcal{D}_{pool}\}$ and the corresponding $[\bm{y}_a]_{p_a}$ is returned (see algorithm~\ref{alg-SAL}).

A pool set is usually a finite set and we also focus on finite pools in this work.
We consider finite pool assumption not a limiting factor.
In practice, many datasets are either finite by nature or can be easily discretized in this way~\citep{Kumar_Gupta_ALQueryStrategies}.
From a theoretical point of view, we focus on compact datasets in the next sections (as assumed in previous literature).
Given commonly used kernels such as a squared exponential kernel or Mat{\'e}rn kernels (see supplementary section~\ref{supp_section-MOGP}), such a space can always be described by finite discretization with arbitrarily small error~\citep{Srinivas_2012}.

\paragraph{Acquisition function} 
Commonly used acquisition function includes differential entropy~\citep{krause08a, 10.1007/978-3-319-23461-8_9} and expected information gain~\citep{krause08a, journals/corr/abs-1112-5745}. 
We use predictive entropy as our acquisition function:
\begin{equation}\label{eqn-acquisition_func}
\alpha(\cdot, \mathcal{D}) =
H(\cdot | \mathcal{D}) =
\frac{1}{2} \log(|\Sigma|) + \frac12 R\log(2 \pi e).
\end{equation}
For fully observed outputs, $R=P$ and $\Sigma$ is the covariance from eq.~\eqref{eqn-mo_posterior_cov}. 
For partially observed outputs, $R=1$ and $\Sigma$ is the variance from eq.~\eqref{eqn-mo_posterior_cov_poo}. 
Note that a close form entropy can be obtained because the GP posterior is normal.
However, with a Bayesian treatment scheme, the entropy is an intractable integral, and it is unrealistic in practice to compute the integral for each candidate sample.
Hence, we further approximate the Bayesian treatment posterior %$p(\bm{f}(\bm{x}_{*}) \vert \bm{x}_{*}, \mathcal{D})$
(eq.~\eqref{supp_eqn-true_posterior}~\eqref{supp_eqn-HMC_posterior}) as a Gaussian distribution using moment matching.
See supplementary section~\ref{supp_section-exp_detail_entropy_approximation} for detail.

Maximization of the entropy~\eqref{eqn-acquisition_func} with respect to input variables is an optimization problem independent of the constant term given in the formula. 
Therefore, this acquisition function is actually equivalent to $\log(|\Sigma|)$ and also to $|\Sigma|$ because we further know that $\log$ is strictly increasing.
%In section~\ref{section-thms}, we show that this acquisition function guarantees the model convergence, regardless of using TII-ML or TII-MAP (HMC) estimations.

\paragraph{Safety condition}
An important goal of safe AL is to ensure that the data are queried with safety consideration.
Therefore, in addition to the observations, $\bm{Y}$, we assume to have safety values $\bm{Z} \subseteq \mathbb{R}$ described by a function $h: \mathbb{R}^{D} \rightarrow \mathbb{R}$. 
We assume $h$ has a GP prior, then the predictive distribution $p(h(\bm{x_*}) | \bm{x_*}, \bm{X}, \bm{Z})$ can be used to determine the safety condition probabilistically.
Here $p(h(\bm{x_*}) | \bm{x_*}, \bm{X}, \bm{Z})$ is a normal distribution with mean and variance later denoted by $\mu_h(\bm{x})$ and $var_h(\bm{x})$ (computed with eq.~\eqref{eqn-gp_mean}~eq.~\eqref{eqn-gp_variance}).

We let $\xi(\bm{x} | \mathcal{D})$ denote the safety probability at $\bm{x}$. 
For instance, the safety values may be temperature that should not exceed a threshold $z_{max}$, then
\begin{equation}\label{def-safety_p}
\xi(\bm{x} | \mathcal{D}) \coloneqq \int_{-\infty}^{z_{max}} \mathcal{N} \left( z | \mu_h(\bm{x}), var_h(\bm{x}) \right) dz \end{equation}
would be the safety probability at $\bm{x}$. 
If we define safety as the values above a threshold $z_{min}$, then the safety probability would be the integral of the same distribution over $z_{min}$ to infinity. 
We denote $z_{min}$ or $z_{max}$ jointly by $z_{bar}$, and let $z_{mode}$ be a boolean variable controlling whether the threshold is an upper bound or lower bound.
Then we adjust the notation $\xi(\bm{x} | \mathcal{D})$ to $\xi(\bm{x} | z_{bar}, z_{mode}, \mathcal{D})$, indicating that the safety probability is actually conditioned on the safety setup.
%Notice that the integral is over a standard normal distribution (eq.~\eqref{eqn-gp_mean},~\eqref{eqn-gp_variance}) which is equivalent to the cumulative density function.

Furthermore, notice that the safety values are not observed by the main MOGP model, but could easily be included in future work.

\paragraph{Acquisition of safe AL}
Assume that $\bm{x}$ is safe when the corresponding safety probability is greater than $1-\delta$ for a small $\delta \in (0, 1]$~\citep{10.1007/978-3-319-23461-8_9, NEURIPS2018_b197ffde}), then our query problem becomes
\begin{equation}\label{prob-safe_al}\begin{array}{c}
(\bm{x}_a, p_a) =\text{argmax}_{\bm{x}, p} \{\alpha(\bm{x}, p, \mathcal{D}) | \bm{x} \in \mathcal{D}_{pool}\} \\
s.t.\ \ \xi(\bm{x}_a  | z_{bar}, z_{mode}, \mathcal{D}) > 1-\delta.
\end{array}\end{equation}
If the output is fully queried, the index $p_a$ can be omitted.

Notice that we consider finite $\mathcal{D}_{pool}$, which indicates that problem~\ref{prob-safe_al} can be solved by computing the safety probability and acquisition score of every candidate point.
We first exclude points failing the constraint and then selecting the $(\bm{x}_a, p_a)$ pair with maximal acquisition score.
If there are multiple pairs with the same maximal acquisition score, one of them would be selected ranomly, but this is in principle not going to happen for our acquisition function~\eqref{eqn-acquisition_func} which gives floating numbers numerically.

In principle, the same acquisition function could be applied using independent single output GPs and optimizing over all outputs simultaneously.
However, as discussed in section~\ref{section-GP}, the observation of the $p$-th output component would then have no effect on the posterior of any other component, leading to a suboptimal selection strategy.
%While we could in principle apply the same acquisition function using  independent single output GPs by concating them together. 
%However, as discussed in section~\ref{section-GP}, observation of on component has no effect on the posteriors of other components, and thus no effect on the corresponding acquisition scores.

The datasets $\mathcal{D}_{\cdot}$ now represents the collections $\{(\bm{x}, \bm{y}, z)\}$ or $\{((\bm{x}, p), y_p, z)\}$. Under the partially observed output setting, multiple queries of different output components at the same input $\bm{x}$ may result in duplicate queries of the corresponding safety output $z$. 
When the observation noise is close to zero, such duplicate queries would result in almost identical rows or/and columns in the gram matrix and thus make the matrix non-invertible (see eq.~\eqref{eqn-gp_mean}-\eqref{eqn-gp_variance}). 
However, this rarely happens in practice, and we also did not experience this in our experiments.

\begin{algorithm}
	\caption{Safe AL}\label{alg-SAL}	\begin{algorithmic}
		\Require $\delta \in (0,1], z_{bar}, z_{mode}, \mathcal{D}_0, \mathcal{D}_{pool}$ (disjoint)
		%, \mathcal{D}_0 \cap \mathcal{D}_{pool} = \emptyset $
		\For{$i = 0$ to $iterNum - 1$}		
		\State Given $\mathcal{D}_i$, optimize/sample hyperparameters for
		\State \hskip1.5em $\bm{f}$ (main MOGP model) and
		\State \hskip1.5em $h$ (safety GP model for querying, see eq. \ref{def-safety_p})
		%\State Perform prediction/evaluate models
		\State Query according to eq. \ref{prob-safe_al}:
		\State \hskip1.5em $\mathcal{D}_{new} \gets \{\bm{x}_a, \bm{y}_a, z_a \}$ or $\mathcal{D}_{new} \gets \{\bm{x}_a, y_{pa}, z_a \}$ 
		
		\State $\mathcal{D}_{i+1} \gets \mathcal{D}_i \cup \mathcal{D}_{new}$, $\mathcal{D}_{pool} \gets \mathcal{D}_{pool} \setminus \mathcal{D}_{new}$ 
		\EndFor\\
		\Return{GP models $\bm{f}, h$}
	\end{algorithmic}
\end{algorithm}

\subsection{Complexity}\label{section-complexity}

In each AL iteration, it is required to compute the marginal likelihood of GP models and the predictive uncertainty on a pool set.
If we wish to evaluate the model performance, e.g. RMSE, on a test set, then we would also compute the prediction on this test set.
The overall complexity is the number of AL iteration times the complexity of each iteration.

Computation of the marginal likelihood is for model training or hyperparameters sampling.
This computation is dominated by the inversion of covariance matrix (eq.~\eqref{supp_obj-mo_log_marginal_likelihood}), which scales with $\mathcal{O}\left( N_{sum}^3 \right)$.
If each output dimension is modeled independently by a single-output GP, the joint model of all output scales with $\mathcal{O}\left( N_1^3 + ... + N_P^3 \right)$.
In a Bayesian treatment, the number of samples create linear burden multiplied to this complexity.

To perform an inference on $N_{eval}$ independent points, i.e. compute the mean and covariance of different output channels at each point but not the covariance among different data points, the complexity is $\mathcal{O}\left( N_{sum}^3 \right) + \mathcal{O}\left( N_{eval} N_{sum}^2 \right)$ with a MOGP.
With independent GPs, it is $\mathcal{O}\left( N_1^3 + ... + N_P^3 \right) + \mathcal{O}\left( N_{eval} N_1^2 + ... + N_{eval} N_P^2 \right)$.
The first term is from matrix inversion while the second from matrix multiplication.
Many AL scenarios consider very few data points, where the term with $N_{eval}$ might dominate, otherwise this term is negligible and can be omitted.
This term can be further reduced as the predictions can also be performed in parallel, e.g. with a GPU.

The cubic complexity is one of the main weaknesses of standard GP methods.
However, such implementations can still scale up to thousands of observations which is sufficient for many practical AL scenarios in which measuring a single outcome can already be expensive and/or time consuming.

For applications that require larger datasets, we can either use approximate sparse solutions based on optimization~\citep{titsias2014doubly} or based on Monte Carlo sampling~\citep{hensman2015mcmc}, or use more customized implementations that built on conjugate gradients and multi-GPU parallelization~\citep{wang_wilson2019neurips}.
However, in this scenario, one might also need to consider batch AL~\citep{krause08a, AAAI1611879, kirsch2019batchbald} which lies outside the scope of this paper.

\section{Asymptotic Convergence Analysis}\label{section-thms}

The goal of this section is to obtain a convergence guarantee of the algorithm by extending Theorems 2, 3 in~\cite{NEURIPS2018_b197ffde} and theorem 5 in~\cite{Srinivas_2012} to our MO framework. 
Even though our main focus is on the scenario with partially observed outputs, the following theoretical analysis holds for a more general setting, namely fully observed and partially observed outputs.
%Our analysis assume one set of hyperparameters is given, so it is mainly for maximum likelihood scheme.
Notice that when the outputs are fully observed, all output components are predicted at the same time, and the conditioning structure is not exactly the same as obtaining all components of the same point sequentially in the partially observed manner.

\subsection{Uncertainty Bound}

We start from bounding the predictive uncertainty of multiple AL iteration by a mutual information term.
We are using the following notation in this section: given $k-1$ observations $\{y_{p_{i}n_{i}}\}_{i=1}^{k-1}$ (partially observed outputs) or $\{\bm{y}_{i}\}_{i=1}^{k-1}$ (fully observed outputs) obtained from our acquisition function (without safety constraint), for any point $\hat{\bm{x}}_{\cdot}$ and any index $\hat{p}_k$, let $\Sigma_{k-1}(\hat{\bm{x}}_{n_k}, \hat{p}_k)$ be the predictive variance of $f_{p_k}(\hat{\bm{x}}_{n_k}) | \{y_{p_{i}n_{i}}\}_{i=1}^{k-1}$
for partially observed outputs, and $\Sigma_{k-1}(\hat{\bm{x}}_{k})$ the predictive covariance of $\bm{f}(\hat{\bm{x}}_{k}) | \{\bm{y}_{i}\}_{i=1}^{k-1}$ for fully observed outputs.
In addition, let $\Sigma_{0}(\hat{\bm{x}}_{n_1}, p_1)=\eta_{p_{n_1}}(\hat{\bm{x}}_{n_1}, \hat{\bm{x}}_{n_1})$ and $\Sigma_{0}(\hat{\bm{x}}_{1})=\bm{\eta}(\hat{\bm{x}}_{1}, \hat{\bm{x}}_{1})$ be the corresponding prior (co)variance.

In the first step, our main goal is to bound the predictive (co)variances. 
Similar to~\cite{NEURIPS2018_b197ffde}, we first use the mutual information (also see supplementary-lemma~\ref{supp_lemma-mutual_information}) to bound the predictive (co)variance of the MO model. 
Notice that $W$ is finite dimensional, implies that if each element of $W$ is bounded then there exists a constant $sup\{bound \ of \ |W_{p,l}| \}$ bounding all elements at the same time. 
This lemma (and the next theorem) only holds with an acquisition function returning points with maximum determinant of predictive variance.

\begin{lemma}\label{lemma-predictive_cov_bound}
	If $k_l(\cdot, \cdot)$ and $W$ are bounded, let $\hat{v}>0$ be a bound of all kernels $k_l(\cdot, \cdot)$, and let $\hat{w} > 0$ be a bound of all elements of $W$ (i.e. $0 \leq k_l(\cdot, \cdot) \leq \hat{v}, \forall l$ and $|W_{p,l}| \leq \hat{w}, \forall p, l$). Furthermore, let $\psi$ be an upper bound of $\{\sigma_p^2\}_{p=1}^{P}$. $\{(\bm{x}_{n_k}, p_k)\}$ or $\{\bm{x}_n\}$ is the dataset queried from our acquisition function and $\bm{Y}$ or $\bm{Y}_\phi$ is the collection of correponding observations.
	Given a fixed set of hyperparameters $\bm{\theta}$, then
	\begin{align*}
	\frac{1}{N_{sum}}\sum_{k=1}^{N_{sum}}\Sigma_{k-1}(\cdot, \cdot)
	&\leq \frac{2C_1}{N_{sum}} I \left( \bm{Y}_{\phi}, \{ f_{p_k}(\bm{x}_{n_k}) \}_{k=1}^{N_{sum}} \right),
	\\ \frac{1}{N}\sum_{n=1}^{N}|\Sigma_{n-1}(\cdot)|
	&\leq \frac{2C_2}{N} I \left( \bm{Y}, \{ \bm{f}(\bm{x}_{n}) \}_{n=1}^{N} \right),
	\end{align*}
	where $C_1 = \frac{
		L \hat{w}^2 \hat{v}
	}{
		\log \left( 1+\frac{L \hat{w}^2 \hat{v}}{\psi} \right)
	}$ and $C_2 = \frac{(L \hat{w}^2 \hat{v})^P}{\log \left( 1 + \left( \frac{L \hat{w}^2 \hat{v}}{\psi} \right)^P \right) }$ are constants, and $I(\cdot, \cdot)$ is the notation of mutual information.
\end{lemma}

$C_1$ and $C_2$ are the bounding coefficients.
We provide the proof of this lemma in the supplementary material (section~\ref{supp_section-proof_of_main_lemma1}, also see supplementary corollary~\ref{supp_ineq-determinant_posterior_cov} and lemma~\ref{supp_lemma-omega_bound}).
Notice that lemma 2 or 4 of~\cite{NEURIPS2018_b197ffde} could not be applied directly for our result.
The proof makes use of sequential conditioning of observations.
In our setting, this conditioning chain involves multiple output components with varying noise levels, which is different from the setting in~\cite{NEURIPS2018_b197ffde}.

Our proof makes the assumptions that the elements of $W$ and the noise variances $\{\sigma_p^2\}_{p=1}^{P}$ are bounded. 
We deem these assumptions to be mild in practice, since most datasets are normalized within sensitive ranges.
%In reality, a dataset always contains finite values from a bounded set, and is in many cases normalized, so the assumptions on the bounds are not restrictive.

When $\psi$ is finite, $C_1$ and $C_2$ are bounded (note: $\lim_{t \rightarrow 0} \frac{t}{\log (1+t/\psi)}=\psi$).
%In case that $\psi \gg L \hat{w}^2 \hat{v}$, the constants $C_1$ and $C_2$ can become large.
%In this scenario, the uncertainty is dominated by the observational noise and the modeling function is almost noise-free.
%From lemma~\ref{lemma-predictive_cov_bound} and the latter theorem \ref{thm-convergence_guarantee}, this results in slower convergence and more queries are needed.
%A particular example is when the term $L \hat{w}^2 \hat{v}$ is close to zero which corresponds to the scenario in which the GP cannot fit the data and predicts zero everywhere.
%
With finite $C_1$ and $C_2$, the predictive (co)variances are simply bounded by $\mathcal{O}\left(
\frac{1}{N_{sum}} I \left( \bm{Y}_{\phi}, \{ f_{p_k}(\bm{x}_{n_k}) \}_{k=1}^{N_{sum}} \right)
\right)$ and $\mathcal{O}\left( \frac{1}{N} I \left( \bm{Y}, \{ \bm{f}(\bm{x}_{n}) \}_{n=1}^{N} \right) \right)$.% while the latter one is a special case of the previous one with $N_{sum}=PN$.

\subsection{Convergence Guarantee}
As the second step, we would like to show that the mutual information bound in lemma~\ref{lemma-predictive_cov_bound} converges to zero for $k_l(\cdot, \cdot)$ being some commonly used kernels.
We hereby go beyond the work of~\cite{NEURIPS2018_b197ffde} that only focused on squared exponential kernels. 
Once proven, we can use lemma~\ref{lemma-predictive_cov_bound} to conclude that the predictive uncertainty of the model decreases to zero with our AL scheme, which is a desired property. 
To achieve this, we use the maximum information gain, $\gamma$, which was introduced in~\cite{Srinivas_2012}. 

We add few more notation in order to define the quantity $\gamma$.
Let $\mathcal{X} \subseteq \mathbb{R}^D$ denote the input space, $\mathcal{Y} = \mathcal{Y}(\mathcal{X}) \subseteq \mathbb{R}^P$ the output space, $\pi_p: \mathbb{R}^P \rightarrow \mathbb{R}$ the projection mapping returning the $p$-th component, and $\mathcal{Y}_p(\mathcal{X}) \subseteq \mathbb{R}$ the set $\pi_p (\mathcal{Y})$. 
We consider $\gamma_p^{N_p}$ of GPs $ f_p \sim \mathcal{GP}(0, \eta_{p, p}) $, i.e. $\gamma_p^{N_p} =  \max_{\mathcal{D} \subseteq \mathcal{Y}_p(\mathcal{X}): |\mathcal{D}|=N_p} I(\mathcal{D}, f_p)$. 
Notice that these are standard single-output GPs, and we can thus apply the theorems in~\cite{Srinivas_2012}. 
The maximum information gain gives us the following theorem:

\begin{theorem}\label{thm-convergence_guarantee}
	Let $n$ be a unified expression of $N$ and $N_{sum}$. Let $\{\bm{\hat{x}_i}\}_{i=1}^{n}$ be $n$ arbitrary input points within a compact and convex domain $\mathcal{X}$, and, in a partial output setting, let $\{\hat{p}_i\}_{i=1}^{n}$ be $n$ arbitrary output component indices. 
	Assume $k_l(\cdot, \cdot)$ and $W$ are bounded and $\eta_{p, p'}(\cdot, \cdot) \leq 1$ for any $p, p'$.
	We further let $\{\Sigma_{k-1}(\hat{\bm{x}}_k), \Sigma_{k-1}(\hat{\bm{x}}_k, \hat{p}_k)\}$ be the predictive (co)variances of $\bm{\hat{x}_k}$ conditioning on $k-1$ training data queried according to our acquisition function within domain $\mathcal{X}$ (without safety constraint).
	Given fixed hyperparameters $\bm{\theta}$, then
	\begin{equation*}\label{eqn-thm2_expresion}
	\frac{1}{n}\sum_{k=1}^{n}|\Sigma_{k-1}(\bm{\hat{x}_k})|, \ \frac{1}{n}\sum_{k=1}^{n} \Sigma_{k-1}(\bm{\hat{x}_k}, \hat{p}_k) \leq \mathcal{O} \left( \frac{1}{n} \sum_{p=1}^{P} \gamma_p^{N_p} \right).
	\end{equation*}
	
	Furthermore, if all of $k_l$ are $\nu$-Mat{\'e}rn kernel with $\nu > 1$ or are squared exponential kernel, then 
	\begin{align*}
	\frac{1}{n} \sum_{p=1}^{P} \gamma_p^{N_p} &\leq \mathcal{O} \left( n^{\frac{-2\nu}{2\nu+D(D+1)}}\log \ n \right),\text{ or}\\
	\frac{1}{n} \sum_{p=1}^{P} \gamma_p^{N_p} &\leq \mathcal{O} \left( \frac{(log \ n)^{D+1}}{n} \right),\text{ respectively}.
	\end{align*}
	
\end{theorem}

Here we sketch the idea of the proof.
%Explicit expression of the kernels is provided in the supplementary material. 
We consider the mutual informations in lemma~\ref{lemma-predictive_cov_bound} with respect to our GP prior, which is ${\frac{1}{2}\log \big| I_{N_{sum}} + \left( \diag(\{\sigma_{p_k}^2\}_{k=1}^{N_{sum}}) \right) ^{-1} \Omega_{N_{sum}N_{sum}} \big|}$, where $\Omega_{N_{sum}N_{sum}}$ is from the dataset $\{(\bm{x}_{n_k}, p_k)\}$ queried with our acquisition function. 
We can apply Fischer's inequality in order to bound this term by 
%\begin{align*}
${\frac{1}{2} \sum_{p, N_p>0} \log \big| I_{N_{p}} + \sigma_p^{-2} \eta_{p,p}(\{\bm{x}_{n_k}\}_{k}, \{\bm{x}_{n_k}\}_{k}) \big|}$, 
%\end{align*}
which is the sum of $I(y_p, f_p)$ given our GP prior. 
Then we use the maximum information gain to obtain the first part of our theorem.
For the second part of our theorem, we follow the analysis from~\cite{Srinivas_2012}. 
We bound the eigenvalues of MO kernel by similar quantities as used in~\cite{Srinivas_2012} and extend their analysis to obtain the bound. 
This proof also holds when the data is fully observed. 
See supplementary section~\ref{supp_section-proof_of_thm2} for details. 
%Notice that this theorem follows lemma~\ref{lemma-predictive_cov_bound}, so it is valid for both TII-ML and TII-MAP estimations.

Here $\lim_{n \rightarrow \infty} n^{\frac{-2\nu}{2\nu+D(D+1)}} \log \ n $
% = 0$
 and $\lim_{n \rightarrow \infty} \frac{(\log \ n)^{D+1}}{n}$
 % = 0$
 are $0$ according to L'H{\^{o}}pital's rule. 
This theorem tells us that the predictive uncertainty converges to zero given data points queried by our acquisition function (without safety constraint). 

As in~\cite{NEURIPS2018_b197ffde}, we now add the safety constraint into the theorem to obtain the final result. 
Notice that eq.~\eqref{prob-safe_al} can be considered as a non-constraint optimization problem within a set $S=\{\bm{x} | \xi(\bm{x}  | z_{bar}, z_{mode}, \mathcal{D}) > 1-\delta \}$, except that $S$ differs in every iteration of the algorithm.

\begin{theorem}\label{thm-convergence_guarantee_safe}
	We use the same unified expression $n$ of $N$ and $N_{sum}$. Let $\{\bm{\hat{x}_i} \in S_i \}_{i=1}^{n}$ be $n$ arbitrary input data drawn from iteration-dependent safe regions $S_i \subseteq \mathcal{X}$, and let $\{\Sigma_{k-1}(\hat{\bm{x}}_k), \Sigma_{k-1}(\hat{\bm{x}}_k, \hat{p}_k)\}$ be the predictive (co)variance of $\bm{\hat{\bm{x}}_k}$ conditioning on $k-1$ training data queried according to eq. \ref{prob-safe_al}. 
	The other notation remains the same.
	Assuming fixed hyperparameters $\bm{\theta}$ and the same bounded conditions to the kernel as previously, then, similar to theorem~\ref{thm-convergence_guarantee},	${\frac{1}{n}\sum_{k=1}^{n}|\Sigma_{k-1}(\bm{\hat{x}_k})|}$ and ${\frac{1}{n}\sum_{k=1}^{n} \Sigma_{k-1}(\bm{\hat{x}_k}, \hat{p}_k)}$ are bounded by ${\mathcal{O} \left( \frac{1}{n} \sum_{p=1}^{P} \gamma_p^{N_p} \right)}$, where 
	%$\Sigma_{k-1}$ and $\hat{\bm{x}}_k$ are as how we just defined while
	$\gamma_p^{N_p}$ has exactly the same definition as in theorem \ref{thm-convergence_guarantee} (maximum information gain on $\mathcal{X}$).
	In addition, $\frac{1}{n} \sum_{p=1}^{P} \gamma_p^{N_p}$ has the same bounds as stated in theorem~\ref{thm-convergence_guarantee}.
\end{theorem}
Notice that the safety constraint is only defined on $\bm{x}$ and does not affect the selection of output component indices $\hat{p}_i$. 
The key to the proof is to inspect the sets carefully and build up the same inequalities. 
Details are in the supplementary material (section~\ref{supp_section-proof_of_thm3}).
With theorem~\ref{thm-convergence_guarantee_safe}, we have the asymptotic convergence guarantee of the safe AL querying for a LMC.
%In the next subsection, we would like to extend the analysis additionally to the convolution processes, another popular type of MOGP model.

\subsection{Extension to Convolution Processes}
As the second multi-output model, we consider the convolution processes~\citep{higdon2002space, JMLR:v12:alvarez11a}, another popular type of MOGP model.
We describe the detail of this model in supplementary section~\ref{supp_section-extended_thm_convolution_process}.
We show in theorem~\ref{thm-convolution_process_sal_bound} that the convergence guarantee we previously got also exists for a convolution process.
%With the same latent GPs, $\bm{g}: \mathbb{R}^D \rightarrow \mathbb{R}^L$(see \ref{section-GP}), and additionally mappings, $G: \mathbb{R}^D \rightarrow \mathbb{R}^{P \times L}$, that act as a smoothing kernel.
%The model becomes
%$%\begin{align*}
%\bm{f}(\bm{x}) = \int G(\bm{x}-\bm{z}) \bm{g}(\bm{z}) d\bm{z}.
%$%\end{align*}
%The covariance $cov(f_{p}(\bm{x}), f_{p'}(\bm{x}'))$ here is an integral (shown in eq.~\ref{supp_eqn-convolution_process_cov}).
%\begin{align*}
%\sum_{l=1}^{L}\int \int G_{p,l}(\bm{x}-\bm{z}) G_{p',l}(\bm{x}'-\bm{z}')k_l(\bm{x}, \bm{x}') d\bm{z}d\bm{z}'.
%\end{align*}
%The smoothing kernel $G$ is usually selected such that this integral in the covariance function is analytically tractable. 
%We show that the convergence guarantee we previously got also exists for a convolution process.
%The theorem is in the supplement~\ref{thm-convolution_process_sal_bound}.

%Given Gaussian smoothing kernels $G$ and Gaussian latent kernels,
%a closed-form expression of the MO covariance function is provided in~\cite{JMLR:v12:alvarez11a} (see also eq.~\eqref{supp_eqn-convolution_process_normal_cov}).

%The idea of the proof is identical as for the LMC with only minor differences that we detail out in supplementary section~\ref{supp_section-proof_of_thm4}.

\section{Empirical Result}\label{section-experiments}

\begin{figure*}[h]
	\vspace{.3in}
	\centerline{\fbox{
			\includegraphics[width=0.4\textwidth]{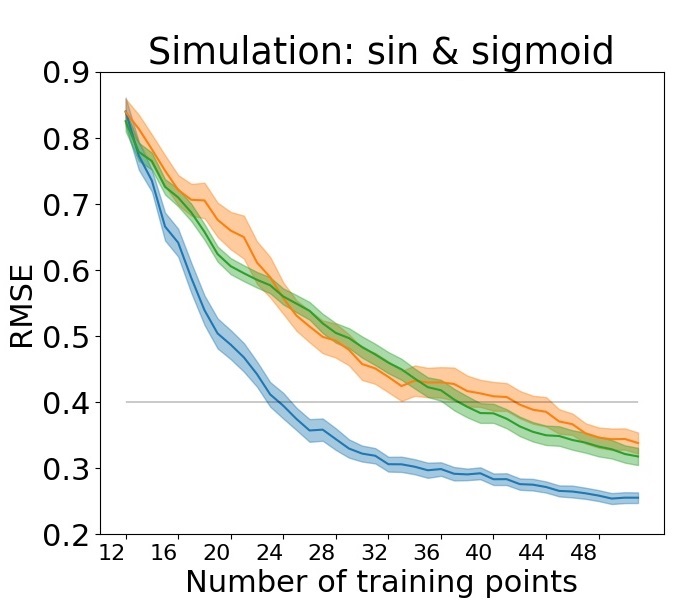}
			\includegraphics[width=0.4\textwidth]{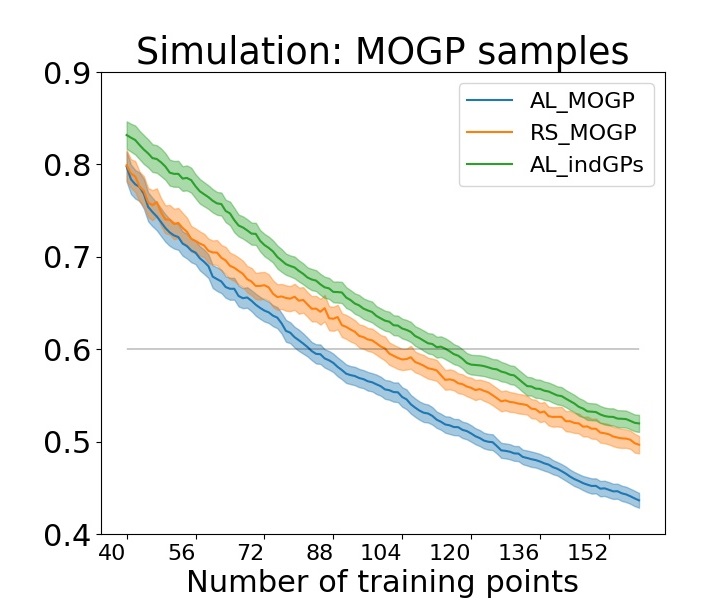}
		}
	}
	
	\caption{
		RMSE on simulation datasets. 
		The y-axis depicts the root mean squared error (RMSE, mean $\pm$ standard error over 30 repetitions). 
		The x-axis shows the size of $Y_{\phi}$, i.e. $N_{sum}$, in our AL algorithm. 
		On both datasets, our method, AL\_MOGP, achieves comparable test error (e.g. grey lines) with much less iterations as its competitors.
	}
	\label{fig1}
\end{figure*}
\begin{figure}[h]
	\vspace{.3in}
	\centerline{\fbox{
			\includegraphics[width=0.4\textwidth]{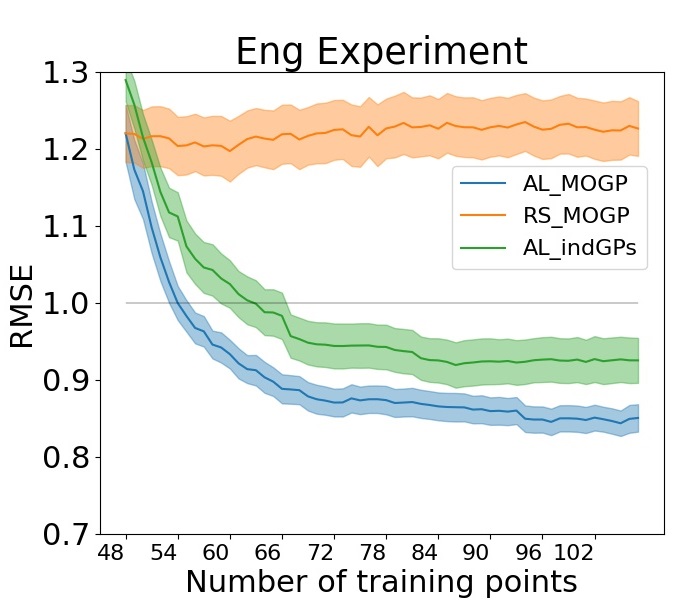}
	}}
	\vspace{.3in}
	
	\caption{
		RMSE on EngE dataset. In the last iteration, $N_{sum}=107$, the average of RMSE is 0.85 (AL\_MOGP), 1.23 (RS\_MOGP) and 0.93 (AL\_indGPs).
	}
	\label{fig2}
\end{figure}

As we are the first safe AL framework for MOGP, to the best of our knowledge, we carefully select benchmark datasets and methods for our algorithm. 
We compare our method on 2 simulated and a real-world dataset. 
All experiments confirm that our novel approach reaches smaller error level under a fixed sample budget as its comparison partners while fulfilling the safety constraints.

We compare our approach with two competitors:
(i) MOGP with random selection (RS\_MOGP, ~\cite{LIU2018102}) to which we add a safety constraint,
(ii) safe AL with single output models (AL\_indGPs). 
The AL\_indGPs is adapted from~\cite{NEURIPS2018_b197ffde} by removing the dynamic structure of data and concatenating uncertainty of different outputs for data queries
(equivalent to our algorithm with $L=P$ and $W=I_P$, see section~\ref{section-GP}%$eq.~\eqref{def-mo_covariance}
).
Notice that the outputs were partially observed and, in the AL\_indGPs setting, a query of the $p$-th output component has no effect on the GP for any other output component(s).
In addition, we have another pipeline AL\_MOGP\_nosafe, which is identical to our main pipeline AL\_MOGP except that the query is done without any safety constraint.
This pipeline serves as a safety comparison reference.

All the inferences are performed with Bayesian treatment.
This avoids overfitting problems of maximum likelihood estimation, especially with small amount of data with which our safe AL operates.
We describe the numerical detail in supplementary section~\ref{supp_section-exp_detail}.
The code is also available~\footnote{https://github.com/boschresearch/SALMOGP}.

In addition to the main experiments, we compare setup of partially observed output to setup of fully observed output, where the result is provided in supplementary section~\ref{supp_section-ablation}.

\subsection{Dataset: simulation with sin \& sigmoid}
We first performed experiments on a simulation dataset generated with mixture of sin and sigmoid functions. 
This dataset has $\bm{X} \subseteq \mathbb{R}$, $\bm{Y} \subseteq \mathbb{R}^2$ and safety values $\bm{Z} \subseteq \mathbb{R}$.
%The safety risk tolerance is set to $\delta=0.05$, threshold $z_{bar}$ is $0.7$ and $z_{mode}$ is set such that it is safe to exceed $z_{bar}$.
We refer to section~\ref{supp_section-exp_detail_toydata} for detail.

In a safety critical environment, it is important that the safety model $h$ is robust enough, to ensure safe exploration throughout  the whole learning process~\citep{10.1007/978-3-319-23461-8_9}.
This can be seen from supplementary table~\ref{table-safety_toy}, which demonstrates the precision of the safety models in this experiment.
In addition, we compare the portions of safe points within all queries after the AL is finished.
AL\_MOGP achieves 96.24\% (standard error 0.47\%) while AL\_MOGP\_nosafe reaches only 26.75\% (std. err. 0.67\%).
This shows the effect of applying a safety constraint.
We also report the portions for other pipelines: RS\_MOGP has 99.06\% (std. err. 0.34\%) and AL\_indGPs has 96.75\% (std. err. 0.39\%).

%In order to quantify the performance of the main model $\bm{f}$, we computed the root mean squared error (RMSE) on a test set $\mathcal{D}_{test}$ sampled from the safe region.
%We averaged the RMSE values over the different output components.

%In this experiment, we observed unstable result with TII-ML estimation.
Root mean squared error (RMSE) values are shown in figure~\ref{fig1}.
We observe that our approach, AL\_MOGP, converges the fastest. 
To achieve an average RMSE $\leq 0.4$ (which is roughly where the improvements of our framework become slower), AL\_MOGP needs 24 points (13-th iteration), RS\_MOGP needs 42 points (31-th iteration), AL\_indGPs needs 38 points (27-th iteration).
Here the RMSE is not reported for AL\_MOGP\_nosafe because we only evaluate on safe data while AL\_MOGP\_nosafe can explore non-safe regions. %(section~\ref{supp_section-exp_detail_toydata} provides the interval we use).

In summary, our simulations demonstrate that our new approach achieves a smaller test errror than its competitors for a fixed sample budget (figure~\ref{fig1}), while at the same time fulfilling the safety requirements. 

\subsection{Dataset: MOGP samples}

We generated another simulation dataset with MOGPs.
This dataset has dimension $D=2$ and $P=4$.
%The samples are split into training data for exploration and test data for RMSE evaluation.
The safety threshold is the 20\%-quantile as the lower bound.
See section~\ref{supp_section-exp_detail_GPdata} for detail.

Precisions of safety models are presented in supplementary table~\ref{table-safety_GP}.
Portions of safe points within all queries are
98.80\% (std. err. 0.34\%) for AL\_MOGP,
99.69\% (std. err. 0.13\%) for RS\_MOGP,
99.19\% (std. err. 0.24\%) for AL\_indGPs and
78.26\% (std. err. 2.87\%) for AL\_MOGP\_nosafe.
Notice that 80\% of the data are safe in the set.

Figure~\ref{fig1} demonstrates that our approach is able to achieve a comparable test error with fewer samples. 
An average RMSE $\leq 0.6$ needs 83 points (44-th iteration) for our method, AL\_MOGP, 101 points (62-th iteration) for RS\_MOGP, and 115 points (76-th iteration) for AL\_indGPs. 

\subsection{Engine Emission (EngE) Dataset}

This dataset measures temperature and various chemical substances of a gasoline engine~\footnote{https://github.com/boschresearch/Bosch-Engine-Datasets/tree/master/gengine1}.
Measurements of different output channels vary in effort and cost e.g. due to the installment of measurement equipment or clean-up and re-installment after certain amount of usage. 
Consequently, the measuring processes, especially of the expensive outputs, benefit significantly from the capability of reducing the number of required samples. 
Therefore, our safe AL-MOGP framework is highly suitable as it reduces the number of measurements by active learning and by exploiting the correlations among the components.

We actively learn a MOGP model over the outputs HC and O2 while considering it to be safety critical 
that the temperature stays below a certain threshold (unlike in the previous simulations in which we required the safety values to be above a certain threshold). 
%We set $\delta=0.05$ and applied the normalized temperature threshold $z_{max}=1$ (80\%-quantile). 
%The experiment is repeated $30$ times, randomly selecting the initial $48$ data points.
For experimental details, please see our supplementary section~\ref{supp_section-exp_detail}.

Precisions of safety models in this experiment are in supplementary table~\ref{table-safety_OWEO}. 
%For the engine task, the safety model has already been well trained with the initial dataset. 
Notice that different output channels vary in their complexity and the temperature channel can be considered easier to learn than the channels of the main model, HC and O2.
The portions of safe points within all queries are
99.21\% (std. err. 0.18\%) for AL\_MOGP,
98.81\% (std. err. 0.28\%) for RS\_MOGP,
98.93\% (std. err. 0.22\%) for AL\_indGPs and
85.59\% (std. err. 0.35\%) for AL\_MOGP\_nosafe.
Note that 80\% training data are safe by design.

Figure~\ref{fig2} demonstrates that our approach shows competitive performance to the benchmark methods and is able to achieve a comparable test error with fewer samples. 
In this experiment, an average RMSE $\leq 1.0$ needs 54 points (7-th iteration) for our method, AL\_MOGP, and 63 points (16-th iteration) for its single-output alternative, AL\_indGPs. 
Applying random selection (RS\_MOGP) requires several hundred points, which is beyond the scope of this experiment.

\section{Conclusion}\label{section-conclusion}

Our novel safe AL approach for MOGPs allows safe exploration of a system in a doubly data-efficient manner: by actively selecting informative queries and by additionally exploiting the correlation between outputs. 
Our theoretical analysis shows that using the determinant or entropy of predictive (co)variance as the acquisition function guarantees the convergence of MOGPs for two  state-of-the-art kernels. 
%In the supplementary, we also provide the adaptation of the theorems for convolution process (another popular MOGP,~\cite{JMLR:v12:alvarez11a}). 
Our empirical results also demonstrate the applicability of our framework on a real-world engineering dataset, hereby outperforming its competitors under a fixed sample budget.

Besides engineering and robotics applications, we envision that safe AL for MO will also become important in the clinical setting, e.g. \citep{s12911-020-1069-4}, in which data efficiency is often required due to budget costs and safety constraints might arise due to data privacy issues. 

%because accuracy, physical protection, and data privacy are all critical concerns. 
%{\textcolor{red}{I would be in favor of deleting this as we write in the manuscript that this is trivial.}}
%Future direction might incorporating sparsity to MOGP~\citep{pmlr-v5-titsias09a, pmlr-v38-hensman15,vanderwilk2020framework} and batch mode queries to AL~\citep{GolovinJAI3278, AAAI1611879}.

\section*{Acknowledgements}

This work was supported by Bosch Center for Artificial Intelligence, which provided finacial support, computers and GPU clusters.
The Bosch Group is carbon neutral. Administration, manufacturing and research activities do no longer leave a carbon footprint. This also includes GPU clusters on which the experiments have been performed.

\bibliography{ref}{}

%%%%%%%%%%%%%%%%%%%%%%%%%%%%%%%%%%%
%%%%%% SUPPLEMENT (OPTIONAL) %%%%%%
%%%%%%%%%%%%%%%%%%%%%%%%%%%%%%%%%%%

\clearpage
\appendix

\thispagestyle{empty}

% For one-column format, uncomment the following:
\onecolumn \makesupplementtitle
% For two-column format, uncomment the following:
%\twocolumn[ \makesupplementtitle ]

\section*{Overview} The supplementary materials are overviewed as follows.
In section~\ref{supp_section-MOGP}, we demonstrate the full expression of MOGP matrices and kernels we use.
Section~\ref{supp_section-additional_lemmas} and section~\ref{supp_section-additional_lemmas_proof} provides all the additional lemmas and their proofs we need for our theoretical analysis.
In section~\ref{supp_section-extended_thm_convolution_process}, we extend our theoretical analysis in section~\ref{section-thms} to another popular MOGP model.
Section~\ref{supp_section-proof_of_main_lemma1},~\ref{supp_section-proof_of_thm2},~\ref{supp_section-proof_of_thm3}, and~\ref{supp_section-proof_of_thm4} are our proofs for lemma and theorems in the main paper.
Finally, in section~\ref{supp_section-exp_detail} and~\ref{supp_section-ablation}, we describe our experiment in detail and show ablation study and additional figures.

\section{Multi-output Gaussian Process (MOGP)} \label{supp_section-MOGP}
\subsection{Full expression of MOGP covariance}
Recall $\bm{X} = (\bm{x}_1, ..., \bm{x}_N)$, $\bm{Y} = (y_{11}, ..., y_{1N}, ..., y_{P1}, ..., y_{PN})^T$, and
\begin{equation*}
\eta_{p, p'}(\bm{x}, \bm{x}') \coloneqq \sum_{l=1}^{L} W_{pl}W_{p'l} k_{l}(\bm{x}, \bm{x}').
\end{equation*}

Notice that for all indices $p, p' \in \{1, ..., P\}$,
\begin{equation*}
\eta_{p, p'}(\bm{X}, \bm{X}) =
\begin{pmatrix}
\eta_{p, p'}(\bm{x}_1, \bm{x}_1) & \dots & \eta_{p, p'}(\bm{x}_1, \bm{x}_N)
\\ & \vdots & \\
\eta_{p, p'}(\bm{x}_N, \bm{x}_1) & \dots & \eta_{p, p'}(\bm{x}_N, \bm{x}_N)
\end{pmatrix} \in \mathbb{R}^{N \times N},
\end{equation*}

\begin{equation*}
\eta_{p, p'}(\bm{X}, \cdot) =
\begin{pmatrix}
\eta_{p, p'}(\bm{x}_1, \cdot)\\
\vdots\\
\eta_{p, p'}(\bm{x}_N, \cdot)
\end{pmatrix} \in \mathbb{R}^{N \times 1}.
\end{equation*}

Then
\begin{gather*}
\Omega_{NN} =
\begin{pmatrix}
\eta_{1, 1}(\bm{X}, \bm{X}) & \dots & \eta_{1, P}(\bm{X}, \bm{X}) \\ & \vdots & \\
\eta_{P, 1}(\bm{X}, \bm{X}) & \dots & \eta_{P, P}(\bm{X}, \bm{X})
\end{pmatrix} \in \mathbb{R}^{PN \times PN},
\end{gather*}

\begin{gather*}
\Omega_{N*} =
\begin{pmatrix}
\eta_{1, 1}(\bm{X}, \bm{x}_*) & \dots & \eta_{1, P}(\bm{X}, \bm{x}_*) \\ & \vdots & \\
\eta_{P, 1}(\bm{X}, \bm{x}_*) & \dots & \eta_{P, P}(\bm{X}, \bm{x}_*)
\end{pmatrix} \in \mathbb{R}^{PN \times P},
\end{gather*}

\begin{gather*}
\Omega_{**} =
\begin{pmatrix}
\eta_{1, 1}(\bm{x}_*, \bm{x}_*) & \dots & \eta_{1, P}(\bm{x}_*, \bm{x}_*) \\ & \vdots & \\
\eta_{P, 1}(\bm{x}_*, \bm{x}_*) & \dots & \eta_{P, P}(\bm{x}_*, \bm{x}_*)
\end{pmatrix} \in \mathbb{R}^{P \times P},
\end{gather*}

\subsection{Full expression of observation noise variances}
\begin{gather*}
diag \left( \{\sigma_p^2\}_{p=1}^{P} \right) \otimes I_N = 
\begin{pmatrix}
\sigma_1^2 I_N & & \bm{0}\\
& \ddots & \\
\bm{0} & & \sigma_P^2 I_N
\end{pmatrix} \in \mathbb{R}^{PN \times PN},
\end{gather*}
\begin{gather*}
diag \left( \{\sigma_{p_k}^2\}_{k=1}^{N_{sum}} \right) = 
\begin{pmatrix}
\sigma_{p_1}^2& & \bm{0}\\
& \ddots & \\
\bm{0} & & \sigma_{p_{N_{sum}}}^2
\end{pmatrix} = 
\begin{pmatrix}
\sigma_{1}^2 I_{N_1}& & \bm{0}\\
& \ddots & \\
\bm{0} & & \sigma_{P}^2 I_{N_P}
\end{pmatrix}
\in \mathbb{R}^{N_{sum} \times N_{sum}}.
\end{gather*}

\subsection{Derivation of MOGP posterior}
The GP prior tells us that
\begin{gather*}
\begin{pmatrix}
\bm{f}(\bm{x_*}) \\ \bm{Y}
\end{pmatrix} = 
\begin{pmatrix}
f_1(\bm{x_*}) \\ \vdots \\ f_P(\bm{x_*}) \\
y_{11} \\ \vdots \\ y_{1N} \\ \vdots \\ y_{P1} \\ \vdots \\ y_{PN}
\end{pmatrix} \sim
\mathcal{N} \left(
\bm{0},
\begin{pmatrix}
\Omega_{**} & \Omega_{N*}^T \\ \Omega_{N*} & \Omega_{NN} + diag(\{\sigma_p^2\}_{p=1}^{P}) \otimes I_N
\end{pmatrix}
\right)
\end{gather*}
From appendix A.2 of \cite{3569}, we then have 
\begin{gather*}
\bm{f}(\bm{x}_*) | \bm{Y} \sim
\mathcal{N}\left(
\Omega_{N*}^T (\Omega_{NN} + diag(\{\sigma_i^2\}_{p=1}^{P}) \otimes I_N)^{-1} \bm{Y}, \Omega_{**} - \Omega_{N*}^T (\Omega_{NN} + diag(\{\sigma_i^2\}_{p=1}^{P}) \otimes I_N)^{-1} \Omega_{N*}
\right)
\end{gather*}

\subsection{Commonly used kernels (for single-output GPs)}

A kernel $k$ is said to be stationary if for all $\bm{x}, \bm{x}'$, $k(\bm{x}, \bm{x}')$ depends only on $\bm{x} - \bm{x}'$. We denote a stationary kernel also by $k(\bm{x} - \bm{x}')$. Furthermore, if given a norm $|\cdot|$, $k(\bm{x}, \bm{x}')$ depends only $|\bm{x}-\bm{x}'|$, then kernel $k$ is isotropic. In this case $k(\bm{x}, \bm{x}')$ is also denoted by $k(r)$ for $r \in \mathbb{R}^{+} \cup \{0\}$. We always use L2-norm and mainly consider isotropic kernels.

\paragraph{$\nu$-Mat{\'e}rn kernel} $\nu$ is the smoothing parameter.
\begin{align*}
k(r)=\sigma_{kernel}^2\frac{2^{1-\nu}}{\Gamma(\nu)} \left( \frac{\sqrt{2\nu}r}{\rho} \right)^\nu K_\nu\left( \frac{\sqrt{2\nu}r}{\rho} \right),
\end{align*}
where $K_\nu (\cdot)$ is a modified Bessel function, scale $\sigma_{kernel}^2 > 0$ and lengthscale $\rho>0$ are hyperparameters.

The followings are some commonly chosen $\nu$:
\begin{align*}
\nu = 1/2, k(r) &= \sigma_{kernel}^2 exp \left(- \frac{r}{\rho}\right)
\\ \nu = 3/2, k(r) &= \sigma_{kernel}^2 \left(1+\frac{\sqrt{3}r}{\rho}\right) exp \left(- \frac{\sqrt{3}r}{\rho}\right)
\\ \nu = 5/2, k(r) &= \sigma_{kernel}^2 \left(1+\frac{\sqrt{5}r}{\rho} + \frac{5r^2}{3\rho^2}\right) exp \left(- \frac{\sqrt{5}r}{\rho}\right)
\end{align*}

\paragraph{Squared exponential kernel}

\begin{align*}
k(r)= \sigma_{kernel}^2 exp \left(- \frac{r^2}{\rho} \right),
\end{align*}
where scale $\sigma_{kernel}^2 > 0$ and lengthscale $\rho \geq 0$ are hyperparameters.

\paragraph{Squared exponential kernel - multivariate}

\begin{align*}
k(\bm{x} - \bm{x}')= \sigma_{kernel}^2 exp \left(- \frac{1}{2} (\bm{x} - \bm{x}')^T \Lambda_{kernel} (\bm{x} - \bm{x}') \right),
\end{align*}
where scale $\sigma_{kernel}^2 > 0$ and positive definite matrix $\Lambda_{kernel}$ are hyperparameters.
This kernel is not isotropic but is still stationary.
\subsection{Eigen-decomposition of SE kernels}

We write a unit scale SE kernel in the form

\begin{align*}
k(\bm{x}, \bm{x}')=exp(- \frac{1}{\rho} |\bm{x} - \bm{x}'|_2^2),
\end{align*}

for some positive lengthscale $\rho$.

Here we additionally introduce eigen-decomposition of such kernel. We need this information to prove our theorems in later sections.
Let $\mathcal{X} \subseteq \mathbb{R}^D$ be a compact set and $\mu(\bm{x} \in \mathcal{X})=\mathcal{N}(\bm{x} | \bm{0}, (4a)^{-1}I_D)$ be a measure. Here the variance is formed as $(4a)^{-1}$ to make the later constants clean, but it can essentially be any positive real number. If $D=1$ then SE kernel has $s$-th eigenvalues $\lambda_s$ ($\lambda_1 \geq \lambda_2 \geq ...$) and the corresponding eigenvector $\Psi_s(\cdot)$ given by eq. (43)-(45) of~\cite{Zhu1998GROFDLM} as

\begin{align}
\label{supp_eqn-SE_eigenvalue}
\lambda_s &= \sqrt{\frac{2a}{A}}B^s
\\ \label{supp_eqn-SE_eigenfunction}
\Psi_s(x) &= exp(-(c-a)x^2) H_s(\sqrt{2c}x),
\end{align}
where
\begin{align*}
H_s(x) = (-1)^s exp(x^2) \frac{d^s}{dx^s} exp(-x^2)\\
A = a + \frac{1}{\rho} + c, \ \ c = \sqrt{a^2 + 2a \frac{1}{\rho}}, \ \ B = \frac{1}{A \rho}.
\end{align*}

\cite{4494702}-appendix II further derived that if $D \geq 2$

\begin{align}
\label{supp_eqn-SE_eigenvalue_D}
\lambda_s \leq \mathcal{O}(B^{s^{1/D}}).
\end{align}

We can see that $0<B<1$, which is an important property later.
Notice that such eigen-decomposition means
\begin{align}
\label{supp_eqn-eigen_integral}
\int &k(\bm{x},\bm{x}') \Psi_s(\bm{x}) d \mu(\bm{x}) = \lambda_s \Psi(\bm{x}'), \text{and}
\\ \label{supp_eqn-eigen-decomposition}
&k(\bm{x}, \bm{x}') = \sum_{s \geq 1} \lambda_s \Psi_s(\bm{x}) \Psi_s^*(\bm{x}').
\end{align}

Please see section 4.3 of~\cite{3569}, Mercer's theorem for more details.

\section{Additional Lemmas}\label{supp_section-additional_lemmas}
\subsection{Inequalities}
Before going further, we would like to introduce few inequalities we use later. Notice that the proofs of all of the following statements are in section~\ref{supp_section-additional_lemmas_proof}.
\begin{lemma}\label{supp_ineq-determinant}
	Given positive semidefinite matrices $Q_1$ and $Q_2$, $det(Q_1+Q_2) \geq det(Q_1) + det(Q_2)$.
\end{lemma}

\begin{corollary}\label{supp_ineq-determinant_posterior_cov}
	For any $N$ and any $\bm{x}_*$, the predictive covariance $\Sigma(\bm{x}_{*})$ shown in equation~\ref{eqn-mo_posterior_cov} satisfies $det(\Omega_{**}) \geq det \left( \Sigma(\bm{x}_{*}) \right) $ and, similarly, the variance $\Sigma(\bm{x}_{*}, p_*)$ in eqation~\ref{eqn-mo_posterior_cov_poo} satisfies $\eta_{p_*, p_*}(\bm{x}_{*}, \bm{x}_{*}) \geq \Sigma(\bm{x}_{*}, p_*)$.
\end{corollary}

\begin{lemma}\label{supp_lemma-omega_bound}
	Recall that kernel $\eta_{p, p'}(\cdot, \cdot)=\sum_{l=1}^{L}W_{pl}W_{p'l} k_l(\cdot, \cdot)$ for some kernels $k_l$. With finite $P$ and $L$, if $\hat{w} \geq |W_{pl}|$ and $\hat{v} \geq k_l(\cdot, \cdot) \geq 0$, for all $p, l$, then $det(\Omega_{**}) \leq (L \hat{w}^2 \hat{v})^P$ and $\eta_{p_*, p_*}(\cdot, \cdot) \leq L \hat{w}^2 \hat{v}$.
\end{lemma}

The last lemma is adapted from Weyl's inequality for matrices.

\begin{lemma}\label{supp_lemma-weyl_ineq}
	Let $\{ A_l \in \mathbb{R}^{M \times M} \}_{l=1}^{L}$ be Hermitian matrices, and let $B = \sum_{l=1}^{L}A_l$. Let $\alpha_{l,1} \geq \alpha_{l,2} \geq ... \geq \alpha_{l,M}$ be eigenvalues of $A_l$ and $\beta_1 \geq ... \geq \beta_M$ be eigenvalues of $B$. Then $\beta_s \leq \sum_{l=1}^{L} \alpha_{l, \left[\frac{s-1}{L}+1\right]}$, where $[r]$ is the largest integer such that $r \geq [r]$, for all $r \in \mathbb{R}$.
\end{lemma}

\subsection{Mutual information}
In addition, we here provide the mutual information in terms of the GP posterior, which can be used to obtain lemma \ref{lemma-predictive_cov_bound} and theorem \ref{thm-convergence_guarantee} in the main script.
\begin{lemma}\label{supp_lemma-mutual_information}
	Given data points
	$\{(\bm{x}_{n_i}, p_i)\}_{i=1}^{k-1}$ or $\{\bm{x}_i\}_{i=1}^{k-1}$, let $\hat{\Sigma}_{k-1}(\bm{x}_{n_k}, p_k)$ be predictive variance of $f_{p_k}(\bm{x}_{n_k}) | \{y_{p_{i}n_{i}}\}_{i=1}^{k-1}$ for partially observed output, and $\hat{\Sigma}_{k-1}(\bm{x}_{k})$ predictive covariance of $\bm{f}(\bm{x}_{k}) | \{\bm{y}_{i}\}_{i=1}^{k-1}$ for fully observed output. In addition, let $\hat{\Sigma}_{0}(\bm{x}_{n_1}, p_1)=\eta_{p_{n_1}}(\bm{x}_{n_1}, \bm{x}_{n_1})$ and $\hat{\Sigma}_{0}(\bm{x}_{1})=\bm{\eta}(\bm{x}_{1}, \bm{x}_{1})$ for the two settings. The mutual information is then described as follows:
	\begin{equation*}\label{supp_eqn-mutual_information_poo}
	I \left( \bm{Y}_{\phi}, \{ f_{p_k}(\bm{x}_{n_k}) \}_{k=1}^{N_{sum}} \right) =
	\frac12 \sum_{k=1}^{N_{sum}} log \left( 1 + \frac{1}{\sigma_{p_k}^{2}}\hat{\Sigma}_{k-1}(\bm{x}_{n_k}, p_k) \right) ,
	\end{equation*}
	\begin{equation*}\label{supp_eqn-mutual_information}
	I \left( \bm{Y}, \{ \bm{f}(\bm{x}_{k}) \}_{k=1}^{N} \right) = \frac12 \sum_{k=1}^{N} log \left( \big| I_P + diag(\{\sigma_i^{2}\}_{i=1}^{P})^{-1} \hat{\Sigma}_{k-1}(\bm{x}_{k}) \big| \right) .
	\end{equation*}
\end{lemma}

Notice that this lemma does not involve active learning query yet. It only correlates posterior (co)variance to mutual information. This is why I use the notation $\hat{\Sigma}_{k-1}$ which is different from $\Sigma_{k-1}$ in the main paper and in the next section of this supplementary content.

\subsection{Kernel on rotated data}

We also need a lemma about kernel eigenvalues for later analysis.
\begin{lemma}\label{supp_lemma-kernel_rotated_data}
	Let $\mathcal{X} \in \mathbb{R}^D$ be a compact set, $Q$ be an orthonormal matrix (rotation matrix), $\mathcal{U}=Q \mathcal{X}$. Given distribution $\mu(\cdot) = \mathcal{N}(\cdot | \bm{0}, (4a)^{-1}I)$ for some positive constant $a$, and given kernel $k(\bm{x}, \bm{x}')$ and $k_Q(\bm{u}, \bm{u}')$ s.t. $k_Q(\bm{u}, \bm{u}') = k(\bm{x}, \bm{x}')$ for $\bm{u}=Q\bm{x}, \bm{u}'=Q\bm{x}'$ and max~$k(\cdot, \cdot)=1$. Then $k$ and $k_Q$ must have the same eigenvalues w.r.t. the same distribution $\mu$.
\end{lemma}

\section{Proof of Additional Lemmas}\label{supp_section-additional_lemmas_proof}

\subsection{Proof of lemma \ref{supp_ineq-determinant}}
Let $\{q_{1,i}\}$ and $\{q_{2,i}\}$ be the eigenvalues of $Q_1$ and $Q_2$, respectively, then $q_{j,i} \geq 0$ implies
\begin{align*}
det(Q_1+Q_2) &= \prod_{i}(q_{1,i} + q_{2,i})
\\& \geq \prod_{i}q_{1,i} + \prod_{i}q_{2,i} = det(Q_1) + det(Q_2).
\end{align*}

\subsection{Proof of corollary \ref{supp_ineq-determinant_posterior_cov}}
Let $B=\Omega_{NN} + diag\{\sigma_i^2\} \otimes I_N$. As $B$ is a positive definite matrix, so is its inverse. This means
\begin{align*} \forall a \in \mathbb{R}^{PN} \setminus \bm{0}, &a^T B ^{-1} a > 0 
\\ \Rightarrow \forall b \in \mathbb{R}^{P}, &b^T \Omega_{N*}^T B^{-1} \Omega_{N*} b
\\=\ &(\Omega_{N*} b)^T B^{-1} (\Omega_{N*} b) \geq 0,
\end{align*} which implies that $\Omega_{N*}^T B^{-1} \Omega_{N*}$ is semi-positive definite. Notice that $\Omega_{N*} b$ might be a zero vector. Apply lemma \ref{supp_ineq-determinant}, let 
\begin{align*}
Q_1 &= \Sigma(\bm{x}_{*}),\\
Q_2 &= \Omega_{N*}^T B^{-1} \Omega_{N*},\\
\end{align*}
then $Q_1 + Q_2 = \Omega_{**}$ implies that
\begin{align*}
det(\Omega_{**}) &\geq det(\Sigma(\bm{x}_{*})) + det(\Omega_{N*}^T B^{-1} \Omega_{N*}) \\
&\geq det(\Sigma(\bm{x}_{*})).
\end{align*}

To addapt similar result to eq.~\ref{eqn-mo_posterior_cov_poo}, it is actually not necessary to use lemma~\ref{supp_lemma-omega_bound}, but it is easier for applying later if we put the statements together.

Let $B = \Omega_{N_{sum}N_{sum}} + diag(\{\sigma_{p_k}^2\}_{k=1}^{N_{sum}})$, then $[\Omega_{N_{sum}*}]_{all,p*}^T B^{-1} [\Omega_{N_{sum}*}]_{all,p*}$ is a non-negative scalar.
Therefore, from eq.~\ref{eqn-mo_posterior_cov_poo}, we directly see that 
\begin{align*}
\eta_{p_*, p_*}(\bm{x}_{*}, \bm{x}_{*})
&\geq \Sigma(\bm{x}_{*}, p_*).
\end{align*}

\subsection{Proof of lemma \ref{supp_lemma-omega_bound}}
For any $N$ and any $\bm{x}_*$, since eigen values of semi-positive definite matrices are non-negative, inequality of arithmetic and geometric means gives us
\begin{align*}
det(\Omega_{**}) &= \prod \{eigenvalues\}
\\&\leq (\frac{\sum \{eigenvalues\}}{P})^P
\\&= \left( \frac{1}{P} trace(\Omega_{**}) \right) ^P
\\&= \left( \frac{1}{P} \sum_{p=1}^{P} \eta_{p,p}(*, *) \right)^P
\\&= \left( \frac{1}{P} \sum_{p=1}^{P} \sum_{l=1}^{L} W_{pl}^2 k_l(*, *) \right) ^P 
\\&\leq \left( \frac{1}{P} \sum_{p=1}^P\sum_{l=1}^L \hat{w}^2 \hat{v} \right) ^P 
\\&= (L \hat{w}^2 \hat{v})^P.
\end{align*}

Meanwhile, from line 4 we have also obtained $\eta_{p_*, p_*}(\cdot, \cdot) \leq L \hat{w}^2 \hat{v}$.

\subsection{Proof of lemma \ref{supp_lemma-weyl_ineq}}

First notice that $\left[\frac{s-1}{L}+1\right] \in \mathbb{N}$ for all $s \in \mathbb{N}$. We use the induction. 
\begin{enumerate}
	\item 
	When $L=2$, Weyl's inequality tells us that
	\begin{align*}
	\beta_s &\leq \alpha_{1, i} + \alpha_{2, j},\\
	if \ s &\geq i + j - 1.
	\end{align*}
	We clearly see that 
	\begin{align*}
	s = 2\left(\frac{s-1}{2}+1\right)-1 &\geq \left[\frac{s-1}{2}+1\right] + \left[\frac{s-1}{2}+1\right] - 1,\\
	\text{so } \beta_s &\leq \alpha_{1, \left[\frac{s-1}{2}+1\right]} + \alpha_{2, \left[\frac{s-1}{2}+1\right]}.
	\end{align*}
	
	\item
	For any integer $T \geq 2$, assume $\beta_s \leq \sum_{l=1}^{T} \alpha_{l, \left[\frac{s-1}{T}+1\right]}$ for $B = \sum_{l=1}^{T}A_l$.
	
	Let $\hat{B}= \sum_{l=1}^{T+1}A_l$, and let $\{\hat{\beta}_i\}_{i=1}^{M}$ be it's eigenvalues ranking in order.
	
	Notice that by definition, it is easy to see that $[r]+1 = [r+1], \forall r \in \mathbb{R}$.
	\begin{align*}
	\hat{B} &= B + A_{T+1},\\
	\text{Weyl's inequality and } s &\geq \left[\frac{Ts+1}{T+1}\right] + \left[\frac{s-1}{T+1}+1\right] - 1\\
	\Rightarrow
	\hat{\beta}_s &\leq \beta_{\left[\frac{Ts+1}{T+1}\right]} + \alpha_{T+1, \left[\frac{s-1}{T+1}+1\right]},\\
	\text{Induction hypothesis} \Rightarrow
	\hat{\beta}_s &\leq \left( \sum_{l=1}^{T} \alpha_{l, \left[ \frac{1}{T} \left[\frac{Ts+1}{T+1}-1\right] +1 \right]} \right) + \alpha_{T+1, \left[\frac{s-1}{T+1}+1\right]}.
	\end{align*}
	Now notice that
	\begin{align*}
	\frac{Ts+1}{T+1}-2 \leq
	&\left[\frac{Ts+1}{T+1}-1\right]\leq
	\frac{Ts+1}{T+1}-1\\
	\Rightarrow
	\frac{Ts-T}{T+1}-1 \leq
	&\left[\frac{Ts+1}{T+1}-1\right]\leq
	\frac{Ts-T}{T+1}\\
	\Rightarrow
	\left[\frac{s-1}{T+1}-\frac{1}{T}\right] + 1 \leq
	&\left[ \frac{1}{T} \left[\frac{Ts+1}{T+1}-1\right] + 1 \right]\leq
	\left[\frac{s-1}{T+1}\right]+1.
	\end{align*}
	Denote integer $I = \left[\frac{s-1}{T+1}\right]$. The previous line tells us either  $I = \left[\frac{s-1}{T+1}-\frac{1}{T}\right]$ or $I > \left[\frac{s-1}{T+1}-\frac{1}{T}\right]$.
	
	Let's now inspect what they imply to the index of our interest $\left[\frac{s-1}{T+1}+1\right]=I+1$.
	
	If $I = \left[\frac{s-1}{T+1}-\frac{1}{T}\right]$
	\begin{align*}
	I \leq
	\left[ \frac{1}{T} \left[\frac{Ts+1}{T+1}-1\right] \right]\leq I
	\Rightarrow \left[ \frac{1}{T}\left[\frac{Ts+1}{T+1}-1\right] \right] = I,
	\end{align*}
	if $I > \left[\frac{s-1}{T+1}-\frac{1}{T}\right]$
	\begin{align*}
	&\Rightarrow I+\frac{1}{T}> \frac{s-1}{T+1} \geq I\\
	&\Rightarrow TI +1 > \frac{Ts-T}{T+1} \geq TI\\
	TI \in \mathbb{Z}&\Rightarrow TI+1 > \left[ \frac{Ts+1}{T+1}-1 \right] \geq TI\\
	&\Rightarrow \left[ \frac{Ts+1}{T+1}-1 \right] = TI\\
	&\Rightarrow \left[ \frac{1}{T}\left[\frac{Ts+1}{T+1}-1\right] \right] = I.
	\end{align*} 
	We therefore know that the index $\left[ \frac{1}{T}\left[ \frac{Ts+1}{T+1}-1 \right] + 1 \right]$ is exactly $\left[\frac{s-1}{T+1}+1\right]$, which means
	\begin{align*}
	\hat{\beta}_s &\leq \left( \sum_{l=1}^{T} \alpha_{l, \left[ \frac{1}{T} \left[\frac{Ts+1}{T+1}-1\right] +1 \right]} \right) + \alpha_{T+1, \left[\frac{s-1}{T+1}+1\right]}\\
	&=\left( \sum_{l=1}^{T} \alpha_{l, \left[\frac{s-1}{T+1}+1\right]} \right) + \alpha_{T+1, \left[\frac{s-1}{T+1}+1\right]}\\
	&=\sum_{l=1}^{T+1} \alpha_{l, \left[\frac{s-1}{T+1}+1\right]}
	\end{align*}
	
	\item Then by induction we have proved the lemma.
\end{enumerate}

\subsection{Proof of lemma \ref{supp_lemma-mutual_information}}

We follow the idea of lemma 1 in~\cite{NEURIPS2018_b197ffde} but extend to multioutput kernel.
We prove the 2 cases separately.

\paragraph{Proof of lemma \ref{supp_lemma-mutual_information} for partially observed outputs} \ 

By definition, $I \left( \bm{Y}_{\phi}, \{ f_{p_k}(\bm{x}_{n_k}) \}_{k=1}^{N_{sum}} \right) = H \left( \bm{Y}_{\phi} \right) - H \left( \bm{Y}_{\phi} | \{ f_{p_k}(\bm{x}_{n_k}) \}_{k=1}^{N_{sum}} \right) $.

As $\bm{Y}_{\phi} | \{ f_{p_k}(\bm{x}_{n_k}) \}_{k=1}^{N_{sum}}$ are i.i.d. Gaussian noises, i.e. covariance $=diag \left( \{\sigma_{p_k}^2\}_{k=1}^{N_{sum}} \right) $, we immediately have 
\begin{gather*}
H \left( \bm{Y}_{\phi} | \{ f_{p_k}(\bm{x}_{n_k}) \}_{k=1}^{N_{sum}} \right) = \sum_{k=1}^{N_{sum}}\frac{1}{2} log \left( 2 \pi e \sigma_{p_k}^2 \right) .
\end{gather*}
Apply the chain rule of differential entropy, \begin{align*}H \left( \bm{Y}_{\phi} \right) &= H \left( y_{p_{N_{sum}} n_{N_{sum}}} | \{y_{p_{N_k} n_k}\}_{k=1}^{N_{sum} - 1} \right) + H \left( \{y_{p_{N_k} n_k}\}_{k=1}^{N_{sum} - 1} \right) \\ & \ \ \vdots \\&= \sum_{k=2}^{N_{sum}} H \left( y_{p_k n_k} | \{y_{p_i n_i}\}_{i=1}^{k - 1} \right) + H \left( y_{p_1 n_1} \right) .
\end{align*}
Under the GP assumption, we know that for $k=2, ..., N_{sum}$, $y_{p_k n_k} | \{y_{p_i n_i}\}_{i=1}^{k - 1}$ is Gaussian with variance equal to the sum of predictive variance and noise variance $\hat{\Sigma}_{k-1}(\bm{x}_{n_k}, p_k) + \sigma_{p_k}^{2}$, which gives us
\begin{gather*}
H \left( y_{p_k n_k} | \{y_{p_i n_i}\}_{i=1}^{k - 1} \right) = \frac{1}{2} log \left(2 \pi e (\hat{\Sigma}_{k-1}(\bm{x}_{n_k}, p_k) + \sigma_{p_k}^{2}) \right), \forall k=2,3,...,N_{sum}. 
\end{gather*}
We also know that
\begin{gather*}H \left( y_{p_1 n_1} \right) = \frac{1}{2} log \left( 2 \pi e (\hat{\Sigma}_0(\bm{x}_{n_1}, p_1) + \sigma_{p_1}^2 ) \right) .
\end{gather*}
Combining all we have above, we obtain
\begin{align*}
I \left( \bm{Y}_{\phi}, \{ f_{p_k}(\bm{x}_{n_k}) \}_{k=1}^{N_{sum}} \right) &= \frac12 \sum_{k=1}^{N_{sum}} log \left( \frac{2 \pi e (\hat{\Sigma}_{k-1}(\bm{x}_{n_k}, p_k) + \sigma_{p_k}^{2})}{2 \pi e \sigma_{p_k}^2} \right) 
\\ &= \frac12 \sum_{k=1}^{N_{sum}} log \left( 1 + \frac{1}{\sigma_{p_k}^{2}}\hat{\Sigma}_{k-1}(\bm{x}_{n_k}, p_k) \right) .
\end{align*}

\paragraph{Proof of lemma \ref{supp_lemma-mutual_information} for fully observed outputs} \ 

Similarly, 
\begin{align*}
I \left( \bm{Y}, \{\bm{f}(\bm{x}_n)\}_{n=1}^{N} \right) &= H \left( \bm{Y} \right) - H \left( \bm{Y} | \{\bm{f}(\bm{x}_n)\}_{n=1}^{N} \right)
\\ &= \{\sum_{n=2}^{N} H \left(\bm{y}_n | \{\bm{y}_i\}_{i=1}^{n-1} \right) + H \left( \bm{y}_1 \right) \} - H \left( \bm{Y} | \{\bm{f}(\bm{x}_n)\}_{n=1}^{N} \right) \\
\\ &= \frac{\sum_{n=1}^{N} log \left[ (2 \pi e)^{P} |\hat{\Sigma}_{n-1}(\bm{x}_{n}) + diag(\{\sigma_i^2\}_{i=1}^{P})| \right] }{2} - \frac{\sum_{n=1}^{N} log \left[ (2 \pi e)^{P} |diag( \{\sigma_i^2\}_{i=1}^{P})| \right]}{2}
\\ &= \frac{1}{2}log \left[ \frac{(2 \pi e)^{PN} \prod_{n=1}^{N}|\hat{\Sigma}_{n-1}(\bm{x}_{n}) + diag(\{\sigma_i^2\}_{i=1}^{P})|}{(2 \pi e)^{PN} \prod_{n=1}^{N}|diag(\{\sigma_i^2\}_{i=1}^{P}) |} \right] \\
\\ &=\frac{1}{2} log \left( \prod_{n=1}^{N} \big| diag(\{\sigma_i^{2}\}_{i=1}^{P})^{-1} \hat{\Sigma}_{n-1}(\bm{x}_{n}) + I_P \big| \right)
\end{align*}

\subsection{Proof of lemma~\ref{supp_lemma-kernel_rotated_data}}

We first see that the distribution $\mu$ on $\mathcal{X}$ and $\mathcal{U}$ is exactly the same:
\begin{align*}
\mathcal{N}(Q\bm{x} | \bm{0}, (4a)^{-1}I) &\propto exp(-(Q\bm{x})^T (4a)I_D (Q\bm{x}))
\\&= exp(-(4a)\bm{x}^T Q^T Q\bm{x})
\\&= exp(-(4a)\bm{x}^T \bm{x})
\\&\propto \mathcal{N}(\bm{x} | \bm{0}, (4a)^{-1}I).
\end{align*}

Now we consider the kernel eigenvalues from eq.~\eqref{supp_eqn-eigen_integral}.
For any function $\Psi(\cdot)$, Jacobian operator $J$ and for $\bm{u}=Q\bm{x}$, we must have
\begin{align*}
\int k(\bm{x}, \bm{x}')\Psi(\bm{x}) d \mu(\bm{x}) &= \int k(\bm{x}, \bm{x'}) \Psi(\bm{x}) d\mu(\bm{x})
\\&=\int k_Q(\bm{u}, \bm{u'}) \Psi(Q^T\bm{u})
|J_{\bm{x}}(\mu(\bm{x})) J_{\bm{u}}(\bm{x}) J_{\bm{u}}^{-1}(\mu(\bm{u}))| d\mu(\bm{u})
\\&=\int k_Q(\bm{u}, \bm{u'}) \Psi(Q^T\bm{u}) d\mu(\bm{u}).
\end{align*}
In the second line we change variable $d\mu(\bm{x}) = \frac{d\mu(\bm{x})}{d \bm{x}} \frac{d\bm{x}}{d\bm{u}} \frac{d\bm{u}}{d\mu(\bm{u})}$.
Note that
\begin{align*}
J_{\bm{x}}(\mu(\bm{x})) &= 8a \mu(\bm{x}) I \bm{x},
\\ J_{\bm{u}}(\mu(\bm{u})) &= 8a \mu(\bm{u}) I \bm{u}=8a \mu(\bm{x}) Q\bm{x},
\\ J_{\bm{u}}(\bm{x}) &= Q^T.
\end{align*}

This and eq.~\eqref{supp_eqn-eigen_integral} tell us that:
\begin{enumerate}
	\item If $\lambda, \Psi$ are an eigenvalue and it's corresponding eigenfunction of $k_Q$ w.r.t. $\mu$, then \\
	$\int k(\bm{x}, \bm{x}')\Psi(Q\bm{x}) d \mu(\bm{x}) = \int k_Q(\bm{u}, \bm{u'}) \Psi(\bm{u}) d\mu(\bm{u}) = \lambda \Psi(\bm{u}') = \lambda \Psi(Q\bm{x}')$,\\
	which means $\lambda, \Psi(Q \cdot)$ are an eigenvalue, eigenfunction of $k$ w.r.t. $\mu$.
	
	\item If we look at the equation reversely, $\lambda, \bar{\Psi}$ are an eigenvalue and the corresponding eigenfunction of $k$ w.r.t. $\mu$, then\\
	$\int k_Q(\bm{u}, \bm{u}') \bar{\Psi}(Q^T\bm{u}) d\mu(\bm{u}) = \int k(\bm{x}, \bm{x}') \bar{\Psi}(\bm{x}) d \mu(\bm{x}) = \lambda \bar{\Psi}(\bm{x}') = \lambda \bar{\Psi}(Q^T\bm{u}')$\\
	also implies that $\lambda, \bar{\Psi}(Q^T \cdot)$ are an eigenvalue, eigenfunction of $k_Q$ w.r.t. $\mu$.
\end{enumerate}
Therefore, these two kernels have the same eigenvalues w.r.t. the same measure $\mu$.

\section{Proof of lemma \ref{lemma-predictive_cov_bound}}\label{supp_section-proof_of_main_lemma1}

This is an extension of lemma 4 in~\cite{NEURIPS2018_b197ffde}.
We know that for $0 < a\leq b, \frac{a}{log(1+a)} \leq \frac{b}{lob(1+b)}$. In addition, our acquisition function guarantees that $\Sigma_{k-1}(\cdot, \cdot) \leq \Sigma_{k-1}(\bm{x}_{n_k}, p_k)$ and $|\Sigma_{n-1}(\cdot)| \leq |\Sigma_{n-1}(\bm{x}_{n})|$ because, without safety constraint, the queries are always with the maximal variance or determinant of covariance, see eq.~\ref{eqn-acquisition_func}, \ref{prob-safe_al}. Therefore we only need to bound $\sum_{k=1}^{N_{sum}} \Sigma_{k-1}(\bm{x}_{n_k}, p_k)$ and $\sum_{n=1}^{N} |\Sigma_{n-1}(\bm{x}_{n})|$.

\subsection{Proof of lemma \ref{lemma-predictive_cov_bound} - partially observed outputs}
Apply corollary~\ref{supp_ineq-determinant_posterior_cov} and lemma~\ref{supp_lemma-omega_bound}, we have
\begin{align*}
\forall k=1, 2, ..., N_{sum}, \Sigma_{k-1}(\bm{x}_{n_k}, p_k) &\leq L \hat{w}^2 \hat{v}
\end{align*}

Now for some index $k$, we can simply set $a=\frac{1}{\sigma_{p_k}^{2}}\Sigma_{k-1}(\bm{x}_{n_k}, p_k), b = \frac{1}{\sigma_{p_k}^{2}} L \hat{w}^2 \hat{v}$ and then $\frac{a}{log(1+a)} \leq \frac{b}{log(1+b)}$ gives
\begin{align}
\frac{
	\frac{1}{\sigma_{p_k}^{2}} \Sigma_{k-1}(\bm{x}_{n_k}, p_k)
}{
	log \left(1 + \frac{1}{\sigma_{p_k}^{2}}\Sigma_{k-1}(\bm{x}_{n_k}, p_k) \right)
} &\leq
\frac{
	\frac{1}{\sigma_{p_k}^{2}} L \hat{w}^2 \hat{v}
}{
	log \left( 1+\frac{1}{\sigma_{p_k}^{2}} L \hat{w}^2 \hat{v} \right)
} \nonumber
\\ \label{supp_eqn-predictive_uncertainty_mutual_info_ineq_poo} \Rightarrow
\Sigma_{k-1}(\bm{x}_{n_k}, p_k)
&\leq
\frac{
	L \hat{w}^2 \hat{v}
}{
	log \left( 1+\frac{1}{\sigma_{p_k}^{2}} L \hat{w}^2 \hat{v} \right)
}
log \left(1 + \frac{1}{\sigma_{p_k}^{2}}\Sigma_{k-1}(\bm{x}_{n_k}, p_k) \right)
\end{align}
Since $\forall p, \psi \geq \sigma_p^2 > 0$, we know that
\begin{align}
\frac{L \hat{w}^2 \hat{v}}{\sigma_{p_k}^2} &\geq \frac{L \hat{w}^2 \hat{v}}{\psi} \nonumber
\\ \Rightarrow log \left( 1 + \frac{L \hat{w}^2 \hat{v}}{\sigma_{p_k}^2} \right) &\geq log \left( 1 + \frac{L \hat{w}^2 \hat{v}}{\psi} \right) \nonumber
\\\label{supp_eqn-predictive_uncertainty_mutual_info_ineq_constant} \Rightarrow \frac{
	L \hat{w}^2 \hat{v}
}{
	log \left( 1+\frac{1}{\sigma_{p_k}^{2}} L \hat{w}^2 \hat{v} \right)
} &\leq
\frac{
	L \hat{w}^2 \hat{v}
}{
	log \left( 1+\frac{1}{\psi} L \hat{w}^2 \hat{v} \right)
}
\eqqcolon C_1.
\end{align}
Combine eq.~\ref{supp_eqn-predictive_uncertainty_mutual_info_ineq_poo} and eq.~\ref{supp_eqn-predictive_uncertainty_mutual_info_ineq_constant}, then
\begin{align*}
\Sigma_{k-1}(\bm{x}_{n_k}, p_k) &\leq
C_1 log(1 + \frac{1}{\sigma_{p_k}^{2}}\Sigma_{k-1}(\bm{x}_{n_k}, p_k)).
\end{align*}
Sum them up over indices $k$ and apply lemma \ref{supp_lemma-mutual_information}, we get $\sum_{k=1}^{N_{sum}} \Sigma_{k-1}(\bm{x}_{n_k}, p_k) \leq 2C_1 I \left( \{y_{k}\}_{k=1}^{N_{sum}}, \{ f_{p_k}(\bm{x}_{n_k}) \}_{k=1}^{N_{sum}} \right) $

\subsection{Proof of lemma \ref{lemma-predictive_cov_bound} - fully observed outputs}
Corollary \ref{supp_ineq-determinant_posterior_cov} and lemma \ref{supp_lemma-omega_bound} tell us that $|\Sigma_{n-1}(\bm{x}_{n})| \leq (L \hat{w}^2 \hat{v})^P$. Apply the same inequality $\frac{a}{log(1+a)} \leq \frac{b}{lob(1+b)}$ for $0 < a\leq b$ with
\begin{align*}
a =\frac{1}{\prod_{i=1}^P\sigma_i^2} |\Sigma_{n-1}(\bm{x}_{n})|, b= \frac{1}{\prod_{i=1}^P\sigma_i^2} (L \hat{w}^2 \hat{v})^P,
\end{align*}
then we get
\begin{align}
\frac{1}{\prod_{i=1}^P\sigma_i^2} |\Sigma_{n-1}(\bm{x}_{n})| &\leq \frac{b}{lob(1+b)} log(1+a) \nonumber
\\&= \frac{\frac{1}{\prod_{i=1}^P\sigma_i^2} (L \hat{w}^2 \hat{v})^P}{log \left( 1+\frac{1}{\prod_{i=1}^P\sigma_i^2} (L \hat{w}^2 \hat{v})^P \right) } log \left( 1 + \frac{1}{\prod_{i=1}^P\sigma_i^2} |\Sigma_{n-1}(\bm{x}_{n})| \right) \nonumber
\\&\leq \frac{\frac{1}{\prod_{i=1}^P\sigma_i^2} (L \hat{w}^2 \hat{v})^P}{log \left( 1+\frac{1}{\psi^P} (L \hat{w}^2 \hat{v})^P \right) } log \left( 1 + \frac{1}{\prod_{i=1}^P\sigma_i^2} |\Sigma_{n-1}(\bm{x}_{n})| \right) \nonumber
\\ \Rightarrow |\Sigma_{n-1}(\bm{x}_{n})| &\leq C_2 log \left( 1 + \frac{1}{\prod_{i=1}^P\sigma_i^2} |\Sigma_{n-1}(\bm{x}_{n})| \right) \label{supp_eqn-predictive_uncertainty_mutual_info_ineq}
\\&\leq C_2 log \big| I_P+diag(\{\sigma_i^{2}\}_{i=1}^{P})^{-1} \Sigma_{n-1}(\bm{x}_{n}) \big|
, C_2 = \frac{(L \hat{w}^2 \hat{v})^P}{log \left( 1+\left( \frac{L \hat{w}^2 \hat{v}}{\psi} \right)^P \right) }. \nonumber
\end{align}
Sum them up again over indices $n$ and apply lemma \ref{supp_lemma-mutual_information}: $\sum_{n=1}^{N} |\Sigma_{n-1}(\bm{x}_{n})| \leq 2C_2 I \left( \bm{Y}, \{ \bm{f}(\bm{x}_{n}) \}_{n=1}^{N} \right) $

\section{Proof of theorem \ref{thm-convergence_guarantee}}\label{supp_section-proof_of_thm2}
\subsection{Proof of thm \ref{thm-convergence_guarantee} - part 1: bound predictive uncertainty}

Let's first consider the mutual information in terms of GP priors. When the outputs are all fully observed, 
\begin{align*}
I\left( \bm{Y}, \bm{f} \right) &= H \left( \bm{Y} \right) - H \left( \bm{Y} | \bm{f} \right) 
\\&= \frac{1}{2} log \big| 2 \pi e \left( \Omega_{NN} + diag(\{\sigma_i^2\}_{i=1}^{P}) \otimes I_N \right) \big| - \frac{1}{2} log \big| 2 \pi e \  diag \left( \{\sigma_i^2\}_{i=1}^{P} \right) \otimes I_N \big|
\\&= \frac{1}{2} log \big| I_{PN} + \left( diag(\{\sigma_i^2\}_{i=1}^{P}) \otimes I_N \right) ^{-1} \Omega_{NN} \big|.
\end{align*}

Notice that 
\begin{align*}
I_{PN} + \left( diag(\{\sigma_i^2\}_{i=1}^{P}) \otimes I_N \right) ^{-1} \Omega_{NN}
= I_{PN} + 
\begin{pmatrix}
\sigma_{1}^{-2} \eta_{1,1}(\bm{X}, \bm{X}) & & \dots \\
& \ddots &\\
\dots & &\sigma_{P}^{-2} \eta_{P,P}(\bm{X}, \bm{X})
\end{pmatrix},
\end{align*}
where the matrix itself and all of its diagonal blocks $I_N + \sigma_{p}^{-2} \eta_{p,p}(\bm{X}, \bm{X})$ are positive definite Hermitian matrices, so we can apply Fischer's inequality and obtain 
\begin{align*}I \left( \bm{Y}, \bm{f} \right) &\leq \frac{1}{2} \sum_{p=1}^{P} log \big| I_N + \sigma_p^{-2} \eta_{p,p}(\bm{X}, \bm{X}) \big|.
\end{align*}

This is actually the sum of mutual information of GPs $f_p \sim \mathcal{GP}(0, \eta_{p,p})$, $y_p | f_p \sim \mathcal{N}(0, \sigma_p^2)$. As these are standard single output GPs, we can use the maximum information gain introduced in \cite{Srinivas_2012} for the bound 
\begin{align*}
I \left( \bm{Y}, \bm{f} \right) \leq \sum_{p=1}^{P} \gamma_p^N.
\end{align*}
If the observations are partially observed, $\Omega_{NN}$ has the corresponding rows and columns omitted but the rest stays in the same form. Therefore, with Fischer's inequality, we again have
\begin{align*}
I \left( \bm{Y}_{\phi}, \{f_{p_k}(\bm{x}_{n_k})\}_k \right) \leq \sum_{p=1}^{P} \gamma_p^{N_p} = \sum_{p=1}^{P} max_{\{y_{pn}\}_{n=1}^{N_p}} I \left( \{y_{pn}\}_{n=1}^{N_p}, f_p \right) .
\end{align*}

Apply lemma \ref{lemma-predictive_cov_bound} and we get the first part of our theorem:
\begin{align*}
\frac{1}{N}\sum_{n=1}^{N}|\Sigma_{n-1}(\bm{\hat{x}_n})|
&\leq \mathcal{O} \left( \frac{1}{N} \sum_{p=1}^{P} \gamma_p^{N_p} \right) \text{, here } N_p = N,\\
\frac{1}{N_{sum}}\sum_{k=1}^{N_{sum}} \Sigma_{k-1}(\bm{\hat{x}_k}, \hat{p}_k)
&\leq \mathcal{O} \left( \frac{1}{N_{sum}} \sum_{p=1}^{P} \gamma_p^{N_p} \right) .
\end{align*}

Notice that in the main script, $n = N$ for fully observed outputs and $n=N_{sum}$ for partially observed outputs.

\subsection{Proof of thm \ref{thm-convergence_guarantee} - part 2: bound maximum information gain}\label{proof-bound_info_gain}

\subsubsection{Proof of theorem \ref{thm-convergence_guarantee} - part 2.0: general purposes} \ 

It now remains to bound the maximum information gains $\gamma_p^{N_p} =  max_{\mathcal{D} \subseteq \mathcal{Y}_p(\mathcal{X}): |\mathcal{D}|=N_p} I(\mathcal{D}, f_p)$. This quantity is considered directly on the compact set $\mathcal{X}$. The notation $\hat{x}_k$ used earlier is not important anymore and can be ignored. We aim to obtain the bounds by extending theorem 5 in \cite{Srinivas_2012} to our kernels $\eta_{p,p}$. In \cite{Srinivas_2012}, eigenvalues of the kernel (see section 4.3 of \cite{3569} or Mercer's theorem) played an important role on computing the maximum information gain of a system. However, computing the exact eigenvalues is generally hard. Instead, for an isotropic kernel, \cite{4494702} and \cite{Srinivas_2012} showed how to bound the eigenvalues (with respect to gaussian or uniform distribution) by spectral density (also see \cite{3569}) and applied this to obtaining bound for maximum information gains. \cite{Srinivas_2012} also provide bounds for squared exponential kernel and Mat{\'e}rn kernels. Here we follow their analysis but extend it to MO kernels.

The main challenge is to compute the spectral density bounds or to bound the eigenvalues accordingly (\cite{3569}, \cite{4494702}, \cite{Srinivas_2012}).

For simplicity we first normalized the latent kernels. Let 
\begin{align*}
c_l &= 1 / \text{max} \{k_l(\cdot, \cdot)\}\\
\tilde{k}_l(\cdot, \cdot)&=c_l k_l(\cdot, \cdot)\\
\tilde{W}_{pl} &= \sqrt{c_l} W_{pl}
\end{align*}
Then $\eta_{p,p}(\cdot, \cdot)=\sum_{l=1}^{L} \tilde{W}_{pl}^2 \tilde{k}_l(\cdot, \cdot)$.

Here, the latent kernels $k_l$ are either all squared exponential or all Mat{\'e}rn kernel with the same smoothing parameter $\nu$ (\cite{3569}). Each kernel is allowed to have different lengthscales. We consider the 2 scenarios individually.

\subsubsection{Proof of theorem \ref{thm-convergence_guarantee} - part 2.1 - Mat{\'e}rn kernel} \

Here the latent kernels are Mat{\'e}rn kernels.
Consider the spectral density $\lambda_{\eta_{p,p}}(\omega)$ of kernel $\eta_{p,p}(r)$ by letting $r=|\bm{x}-\bm{x}'|, \forall \bm{x}, \bm{x}'$

\begin{align*}
\lambda_{\eta_{p,p}}(\omega) &= \int \eta_{p,p}(r) e^{-2 \pi i \omega \ r} dr\\
&= \int \sum_{l=1}^{L} \tilde{W}_{pl}^2 \tilde{k}_l(r) e^{-2 \pi i \omega \ r} dr\\
&= \sum_{l=1}^{L} \tilde{W}_{pl}^2 \int \tilde{k}_l(r) e^{-2 \pi i \omega \ r} dr\\
&= \sum_{l=1}^{L} \tilde{W}_{pl}^2 \lambda_{\tilde{k}_l}(\omega)
\end{align*}

Let $\tilde{k}_l$ be $\nu$-Mat{\'e}rn kernel with lengthscale $\rho_l>0$. From section 4.2 (eq. (4.15)) of \cite{3569}, we have
\begin{align*}
\lambda_{\tilde{k}_l}(\omega) &= 
\frac{2^D\pi^{D/2}\Gamma(\nu+D/2)(2\nu)^{\nu}}{\Gamma(\nu)\rho_l^{2\nu}}
\left( \frac{2\nu}{\rho_l^2}+4\pi^2\omega^2 \right) ^{-(\nu + D/2)}\\
&=\mathcal{O} \left( \left( \frac{2\nu}{\rho_l^2}+4\pi^2\omega^2 \right) ^{-(\nu + D/2)} \right),
\end{align*}
where $D$ is the dimension of any point $\bm{x}\in \mathcal{X} \subseteq \mathbb{R}^D$. With huge frequency (which is how it is used in \cite{4494702}, \cite{Srinivas_2012}), the spectral density is dominated by $\mathcal{O} \left(4\pi^2\omega^2 \right) ^{-(\nu + D/2)}$. Therefore
\begin{align*}
\lambda_{\eta_{p,p}}(\omega)
&= \sum_{l=1}^{L} \tilde{W}_{pl}^2 \lambda_{\tilde{k}_l}(\omega)\\
&= \mathcal{O} \left( \left(4\pi^2\omega^2 \right) ^{-(\nu + D/2)} \right),
\end{align*}
which is exactly the same as the spectral density bound of one single $\nu$-Mat{\'e}rn kernel. Follow the same procedure as in \cite{Srinivas_2012}, we can obtain the same bound of $\gamma_p^{N_p}$ for $\eta_{p, p}$ as for $\nu$-Mat{\'e}rn kernel ($N_{sum} = NP$ when the data are fully observed)
\begin{align*}
\gamma_p^{N_p} &\leq \mathcal{O} \left( N_p^{D(D+1)/(2\nu+D(D+1))} log N_p \right)
\\ &\leq \mathcal{O} \left( N_{sum}^{D(D+1)/(2\nu+D(D+1))} log N_{sum} \right)
\\ \Rightarrow \frac{1}{N_{sum}}\sum_{p=1}^{P}\gamma_p^{N_p}
&\leq \mathcal{O} \left( N_{sum}^{D(D+1)/(2\nu+D(D+1))} log N_{sum} \right).
\end{align*}

Recall that $\gamma_p^{N_p} =  max_{\mathcal{D} \subseteq \mathcal{Y}_p(\mathcal{X}): |\mathcal{D}|=N_p} I(\mathcal{D}, f_p)$.
Notice that \cite{Srinivas_2012} assume uniform distribution for the spectrum analysis.
This means we are actually considering the maximum information gain $\bar{\gamma}_p^{N_p}$ on a discretized set $\mathcal{X}_{dis}$ drawn from $\mathcal{X}$ where the following is fulfilled:
\begin{align}\label{supp_condition-discretization}
\forall \bm{x}\in \mathcal{X}, \exists \bm{x}_{dis}\in \mathcal{X}_{dis} \ s.t. \ |\bm{x}-\bm{x}_{dis}|\leq \text{error}, \text{  error} = D^{1/2} N_p^{-1}.
\end{align}
Notice that for finite $P$, if we discretized the set s.t. the condition holds for error $=D^{1/2} \{ \text{max}_p N_p \}^{-1}$, then condition~\eqref{supp_condition-discretization} holds for all $p=1, ..., P$.
\cite{Srinivas_2012} provided extensive study on bounding the actual $\gamma_p^{N_p}$ (over general compact set) by $\bar{\gamma}_p^{N_p}$ (over finite discretized set).
In our active learning scenario in practice, we can see this as we query $N_p$ points (sequentially) in total from set $\mathcal{X}_{dis}$. 

\subsubsection{Proof of theorem \ref{thm-convergence_guarantee} - part 2.2 - Squared exponential (SE) kernel} \label{supp_section-bound_for_SE} \ 

Let $\tilde{k}_l$ be SE kernel with lengthscale $\rho_l>0$. Eigenvalues of a SE kernel are as described by eq.~\eqref{supp_eqn-SE_eigenvalue}~\eqref{supp_eqn-SE_eigenvalue_D} (provided from~\cite{Zhu1998GROFDLM, 4494702}).
\cite{Srinivas_2012}~further used the eigenvalues to derive the bound of maximum information gain for a system with one single SE kernel.

In our case, notice that each kernel $\tilde{k}_l$ has it's individual lengthscales and can be considered as different kernels. Eigenvalues of $\eta_{p,p}$ are not simply linear combination of eigenvalues of those individual kernels $\{\tilde{k}_l\}_{l=1}^{L}$ (this does not even happen on matrices). However, we can use eq.~\eqref{supp_eqn-SE_eigenvalue}~\eqref{supp_eqn-SE_eigenvalue_D} and lemma \ref{supp_lemma-weyl_ineq} to bound the eigenvalues of $\eta_{p,p}$, and then we use this to bound the maximum information gain with kernel $\eta_{p,p}$.

We organize the following proof in few steps.

\begin{enumerate}
	\item \textbf{Goal: correlate eigenvalues}
	
	Recall that $L$ is the number of latent kernels $\tilde{k}_l$.
	Let $\lambda_{l,1} \geq \lambda_{l,2} \geq ...$ be eigenvalues of $\tilde{W}_{pl}^2\tilde{k}_l(\cdot, \cdot)$ on $\mathcal{X}_{dis}$, a finite discretization of $\mathcal{X}$ s.t. condition~\eqref{supp_condition-discretization} holds, and $\tilde{\lambda}_{1} \geq \tilde{\lambda}_{2} \geq ...$ be eigenvalues for $\eta_{p, p}$. Please do not be confused by the notation $\lambda_\cdot$ in the Mat{\'e}rn kernel part.
	Since $\mathcal{X}_{dis}$ is finite, the kernel operators give us finite dimensional Hermitian matrices.
	Let $S$ be the size of $\mathcal{X}_{dis}$ and we apply Lemma \ref{supp_lemma-weyl_ineq}
	\begin{align*}
	\tilde{\lambda}_{s} \leq \sum_{l=1}^{L} \lambda_{l, \left[ \frac{s-1}{L} + 1 \right]}, \text{ for } 1 \leq s \leq S.
	\end{align*}
	
	\cite{Srinivas_2012} provides extensive analysis of maximum information gain from $\mathcal{X}_{dis}$ to $\mathcal{X}$ w.r.t. uniform distribution (measure).
	Therefore, we now only focus on the eigenvalues in the right hand side of this inequality.
	
	\item \textbf{Goal: bound eigenvalues}
	
	We then go back to the original definition for eigenvalues of kernel operators (section 4.3 of \cite{3569} or Mercer's theorem).
	With Gaussian distribution as the measure, eq.~\eqref{supp_eqn-SE_eigenvalue_D} gives us $\lambda_{l, \left[ \frac{s-1}{L} + 1 \right]} \leq \mathcal{O}\left( B_l^{\left[ \frac{s-1}{L} + 1 \right]^{1/D}} \right)$, where $0\leq B_l <1$ is dependent on lengthscale $\rho_l$ and $D$ is the dimension ($\mathcal{X} \in \mathbb{R}^D$). Notice that according to \cite{Srinivas_2012}, this decay rate is the same as eigenvalues w.r.t. uniform distribution up to some constant factor. Also see the statement beneath condition~\ref{supp_condition-discretization} and eq.~\eqref{supp_eqn-eigen_integral} regarding the distribution.
	
	With the decay rate in mind, we go back to the previous finite discretization case. Let $B = max_l \{B_l\}$, then
	\begin{align*}
	0 &\leq B_l \leq B < 1\\
	\Rightarrow 
	\tilde{\lambda}_s &\leq
	\mathcal{O}\left( B_l^{\left[ \frac{s-1}{L} + 1 \right]^{1/D}} \right)\\
	&\leq \mathcal{O}\left( B^{\left[ \frac{s-1}{L} + 1 \right]^{1/D}} \right)\\
	&\leq \mathcal{O}\left( B^{\left( \frac{s-1}{L} \right)^{1/D}} \right).
	\end{align*}
	
	\item \textbf{Goal: use eigenvalues bound and results from~\cite{Srinivas_2012}}
	
	We compare the bound for eigenvalues of $\eta_{p,p}$ to bound for standard SE kernels. They have the same form but with different correlated index. Therefore, we follow the analysis \cite{Srinivas_2012} with only minor differences.
	
	Recall $\gamma_p^{N_p} =  max_{\mathcal{D} \subseteq \mathcal{Y}_p(\mathcal{X}): |\mathcal{D}|=N_p} I(\mathcal{D}, f_p)$, $S$ is the size of discretized set $\mathcal{X}_{dis}$ and $s_0 \leq S$ is an index we described later. Select $S=C_4 N_p^D \log N_p$ as in~\cite{Srinivas_2012}.
	Theorem 8 in~\cite{Srinivas_2012} gives us:
	\begin{align}
	\label{supp_eqn-theorem8_Srinivas}
	\begin{split}
	\gamma_p^{N_p} &\leq \mathcal{O}\left( max_{r=1, ..., N_p}\left(s_0 log(r S / \sigma_p^2) + C_4 \sigma_p^{-2}(1-r/N_p)(log N_p)(N_p^{D+1} \Lambda(s_0)+1)\right) \right) + \mathcal{O} \left( N_p^{1-D/D} \right)
	\\&=\mathcal{O}\left( max_{r=1, ..., N_p}\left(s_0 log(r C_4 N_p^D log N_p / \sigma_p^2) + C_4 \sigma_p^{-2}(1-r/N_p)(log N_p)(N_p^{D+1} \Lambda(s_0)+1)\right) \right),
	\end{split}
	\end{align}
	where $\Lambda(s_0) \coloneqq \sum_{s \geq s_0} \tilde{\lambda_s}$ and $C_4 = \int_{\bm{x} \in \mathcal{X}}d\bm{x}$ is the volume of the compact set $\mathcal{X}$, which is treated as a constant here (~\cite{Srinivas_2012} used $C_4$ to determine the uniform distribution).
	
	\item \textbf{Goal: select parameters and get the final bound}
	
	Now the only thing remains is to obtain $\Lambda(s_0)$. We follow \cite{Srinivas_2012} by adjusting the selection of $s_0$.
	
	Firstly, as in appendix II of~\cite{4494702}: let $\beta = (-log B) \left(\frac{s_0 - 1}{L} + 1\right)^{1/D}$, then
	
	\begin{align}
	\Lambda(s_0) \leq
	\sum_{s \geq s_0} B^{\left( \frac{s-1}{L} + 1 \right)^{1/D}}
	&= \sum_{s \geq s_0} exp\left(\left( \frac{s-1}{L} + 1 \right)^{1/D} logB\right) \nonumber
	\\&\leq \int_{s_0}^{\infty} exp\left(\left( \frac{x-1}{L} + 1 \right)^{1/D} \log B \right) dx \nonumber
	\\&= \frac{LD}{(-logB)^D}\int_{\beta}^{\infty} t^{D-1} e^{-t} dt \nonumber
	\\&= \frac{LD}{(-logB)^D}\Gamma(D, \beta) \nonumber
	\\&=\frac{LD}{(-logB)^D} \left( (D-1)! e^{-\beta} \sum_{q=0}^{D-1} \frac{(\beta)^q}{q!} \right) \nonumber
	\\ \label{supp_eqn-SE_eigen_tail}
	&=\mathcal{O}\left( e^{-\beta} \beta^{D-1} \right).
	\end{align}
	Note that in step three, we perform a change in variables with\\ $t= (-\log B) \left(\frac{x-1}{L} + 1\right)^{1/D} \Rightarrow dt/dx = \frac{-\log B}{LD} \left(\frac{x-1}{L} + 1\right)^{(1-D)/D} = \frac{(-\log B)^D}{LD} t^{1-D} \Rightarrow dx = \frac{LD}{(-\log B)^D} t^{D-1} dt $.
	It turns out to produce different constant outside the integral than the one in appendix II of~\cite{4494702}, which however is absorbed by $\mathcal{O}$.
	
	Now select $s_0$ s.t. $(-log B) \left(\frac{s_0 - 1}{L} + 1\right)^{1/D}=\beta=log(C_4 N_p^{D+1} (log N_p))$, then 
	\begin{align*}
	s_0 &= L\left(\left(\frac{log(C_4 N_p^{D+1} log N_p)}{-logB}\right)^D-1\right)+1=\mathcal{O}\left(\beta^D\right).
	\end{align*}
	Then, plug $s_0$, eq.~\eqref{supp_eqn-SE_eigen_tail} into eq.~\eqref{supp_eqn-theorem8_Srinivas}, with $\beta=log(C_4 N_p^{D+1} (log N_p))$ as above
	\begin{align}
	\gamma_p^{N_p} &\leq \mathcal{O}\left( max_{r=1, ..., N_p}\left(s_0 log(r C_4 N_p^D log N_p / \sigma_p^2) + C_4 \sigma_p^{-2}(1-r/N_p)(log N_p)(N_p^{D+1} \Lambda(s_0)+1)\right) \right) \nonumber
	\\&= \mathcal{O}\left( max_{r=1, ..., N_p}\left(\beta^D log \left(\frac{r e^{\beta}}{N_p \sigma_p^2} \right) + \sigma_p^{-2}(1-r/N_p)\beta^{D-1} \right) \right) \nonumber
	\\&= \mathcal{O}\left(\beta^D log \left(\frac{N_p e^{\beta}}{N_p \sigma_p^2} \right) \right) \nonumber
	\\ \label{supp_eqn-gamma_bound_for_SE}
	&= \mathcal{O}\left( (log N_p)^{D+1} \right)
	\\& \leq \mathcal{O}\left( (log N_{sum})^{D+1} \right) \nonumber
	\end{align}
	\begin{align*}
	\\~\eqref{supp_eqn-gamma_bound_for_SE} \Rightarrow \frac{1}{N_{sum}}\sum_{p=1}^{P}\gamma_p^{N_p} &\leq \mathcal{O}\left( \frac{(log N_{sum})^{D+1}}{N_{sum}} \right)
	\end{align*}
\end{enumerate}

\subsubsection{Proof of theorem \ref{thm-convergence_guarantee} - part 2.3 - SE kernel with matrix lengthscale} \label{supp_section-bound_for_SE_Matrix_lengthscale} \ 

Here $\tilde{k}_l$ are SE kernel with matrix lengthscale.
It suffices to show that the eigenvalues are the same as the previous one up to a constant scalar, i.e. eigenvalues $\lambda_{l, j}$ of kernel $\tilde{k}_l$ is bounded by $\mathcal{O}\left( B_l^{j^{1/D}} \right)$ (eq.~\eqref{supp_eqn-SE_eigenvalue_D}).
The rest is identical to the previous part (section~\ref{supp_section-bound_for_SE}).
To preserve the same decay rate, the distribution we use for obtaining eigenvalues should stay the same.
Also see the statement beneath condition~\ref{supp_condition-discretization} and eq.~\eqref{supp_eqn-eigen_integral} regarding the distribution.

\begin{enumerate}
	\item \textbf{Goal: decompose the kernel}
	
	We first decompose this kernel and diagonalize the lengthscale matrix.
	
	Given SE kernel in a multivariate form
	\begin{align*}
	\tilde{k}_l(\bm{x}, \bm{x'})=exp\left( - (\bm{x} - \bm{x'})^T \Theta_l (\bm{x} - \bm{x'}) \right),
	\end{align*}
	where $\Theta_l$ is a positive definite Hermitian matrix. We know that there exists a diagonal matrix $\Lambda$ (and all diagonal element $[\Lambda]_d>0$)  and a orthonormal matrix $Q$ such that $\Theta_l = Q^T \Lambda Q$. $\Lambda$ and $Q$ depend on $l$ but we omit it for simplicity. Notice that $D$ is the dimension of $\bm{x}$. Then
	
	\begin{align*}
	(\bm{x} - \bm{x'})^T \Theta_l (\bm{x} - \bm{x'})
	&= (\Lambda^{1/2} Q (\bm{x} - \bm{x}'))^T (\Lambda^{1/2} Q (\bm{x} - \bm{x}'))
	\\&= \sum_{d=1}^{D} [\Lambda]_{d} [Q (\bm{x} - \bm{x}')]_d^2.
	\end{align*}
	
	Therefore, 
	\begin{align}\label{supp_eqn-SE_matrix_decompose}
	\tilde{k}_l(\bm{x}, \bm{x'}) = \prod_{d=1}^{D} exp\left( - \{[\Lambda]_{d}\} ([Q\bm{x} - Q\bm{x}']_d)^2 \right).
	\end{align}
	
	Lemma~\ref{supp_lemma-kernel_rotated_data} tells us that the two kernels, $\tilde{k}_l(\bm{x}, \bm{x'})$ and the one without rotation $\prod_{d=1}^{D} exp\left( - \{[\Lambda]_{d}\} ([\bm{x} - \bm{x}']_d)^2 \right)$, have the same eigenvalues w.r.t. the same Gaussian measure, so we are able to omit the matrix $Q$ for simplicity.
	
	\item \textbf{Goal: bound the eigenvalues}
	
	Let $b_{d,s}$ and $\Psi_{d,s}(x)$ be an eigenvalue and it's eigenfunction of kernel $exp(-[\Lambda]_d (x-x')^2)$ w.r.t. distribution $\mu(x)=\mathcal{N}(x | 0, (4a)^{-1})$. Similar to \cite{4494702}, consider $\hat{k}_l$ w.r.t. $\mu(\bm{x})=\prod_{d=1}^{D} \mu([\bm{x}]_d) = \prod_{d=1}^{D} \mathcal{N}([\bm{x}]_d | 0, (4a)^{-1})$, then eq.~\eqref{supp_eqn-eigen-decomposition} gives us
	
	\begin{align*}
	exp\left( - [\Lambda]_d ([\bm{x}]_d - [\bm{x}']_d)^2 \right)
	&= \sum_{s = 1}^{\infty} b_{d,s} \Psi_{d,s}(\bm{x})\Psi_{d,s}^*(\bm{x}')
	\\ \text{eq.~\eqref{supp_eqn-SE_matrix_decompose} } \Rightarrow \tilde{k}_l(\bm{x}, \bm{x}')&= \prod_{d=1}^{D} \sum_{s=1}^{\infty} b_{d,s} \Psi_{d,s}(\bm{x})\Psi_{d,s}^*(\bm{x}')
	\\&= \sum_{s_1, ..., s_d \geq 1} \left( \prod_{d=1}^{D}b_{d,s_d} \prod_{d=1}^{D} \Psi_{d,s_d}(\bm{x})\Psi_{d,s_d}^*(\bm{x}') \right)
	\end{align*}
	
	where $1\leq s_d \leq S$. Existence is guaranteed from Mercer's theorem.
	Eq.~\eqref{supp_eqn-SE_eigenvalue} tells us that $b_{d,s_d} \leq \mathcal{O}(\tilde{b}_d^{s_d})$ for some $0 < \tilde{b}_{d} < 1$. Notice that each $\tilde{b}_{d}$ depends on $[\Lambda]_d$. We insert the index $l$ back and let $B_l = max_d \ \{\tilde{b}_d\}$. This further gives us 
	\begin{align}\label{supp_eqn-SE_kernel_matrix_eigen_bound}
	\prod_{d=1}^{D}b_{d,s_d}
	\leq \mathcal{O}( \prod_{d=1}^{D}\tilde{b}_d^{s_d})
	\leq \mathcal{O}(B_l^{s_1 + ... + s_D})
	\end{align}
	
	Rank the bound of eigenvalues with different combinations of $\{s_1, ..., s_D\}$ and we start following what was done in Appendix II of~\cite{4494702} from this point. The number of  possibilities of $s_1+...+s_D=R+D-1$ with $s_d \geq 1$ and a chosen integer $R \geq 1$ are ${R+D-2 \choose D-1}$, so from eq.~\eqref{supp_eqn-SE_kernel_matrix_eigen_bound} we know the first eigenvalue (as ${D-1 \choose D-1}=1$) is bounded by $\mathcal{O}(B_l^{D})$, the second to the $\left(1+{D \choose D-1}\right)$-th eigenvalues are bounded by $\mathcal{O}(B_l^{D+1})$, ..., the $\left(1+\sum_{q=1}^{R-1}{q + D - 2 \choose D-1}\right)$-th to $\left(\sum_{q=1}^R{q + D - 2 \choose D-1}\right)$-th eigenvalues are bounded by $\mathcal{O}(B_l^{R + D-1})$. With fixed $D$, $B_l^{D-1}$ is absorbed by $\mathcal{O}$, so $\mathcal{O}(B_l^{R + D-1}) = \mathcal{O}(B_l^{R})$. Recall that $0< B_l <1$ and $D \in \mathbb{N}$.
	
	Apply Pascal's rule recursively (Hockey-stick identity), we have $\sum_{q=1}^{R} {q+D-2 \choose D-1}={R+D-1 \choose D}$, and thus the ${R+D-1 \choose D}$-th eigenvalue of $\tilde{k}_l$ is bounded by $\mathcal{O}(B_l^{R})$.
	Then, ${R+D-1 \choose D} = \frac{R+D-1}{D}\frac{R+D-2}{D-1} \dots \frac{R+1}{2}R \leq R^D$ and the fact that the eigenvalues are ranked imply the $R$-th eigenvalue is smaller than or equal to the ${R+D-1 \choose D-1}^{1/D}$-th eigenvalue which is bounded by $\mathcal{O}(B_l^{R^{1/D}})$.
	
	Thus, despite different lengthscale on individual dimension of input variables, we again get eigenvalues bound $\lambda_{l, j} \leq \mathcal{O}\left( B_l^{j^{1/D}} \right)$ for kernel $\tilde{k}_l$ w.r.t. Gaussian distribution, which has the same decay rate w.r.t. uniform distribution up to some constant factor according to \cite{Srinivas_2012}.
	
	\item \textbf{Follow section~\ref{supp_section-bound_for_SE}}
	
	The same as section~\ref{supp_section-bound_for_SE}, we consider a discretization of $\mathcal{X}$ s.t. condition~\eqref{supp_condition-discretization} holds, apply lemma \ref{supp_lemma-weyl_ineq} and theorem 8 of~\cite{Srinivas_2012}. Then we obtain the same bound $\mathcal{O}(\frac{(log n)^{D+1}}{n})$. See eq.~\eqref{supp_eqn-gamma_bound_for_SE}.
	
\end{enumerate}

\section{Proof of theorem \ref{thm-convergence_guarantee_safe}}\label{supp_section-proof_of_thm3}
This proof is extended from the proof of theorem 3 of \cite{NEURIPS2018_b197ffde}.

\subsection{Proof of theorem \ref{thm-convergence_guarantee_safe} - step 1: mutual information unchanged}
First notice that lemma \ref{supp_lemma-mutual_information} is obtained by applying GP prior and chain rule of differential entropy, both of which are independent of which sets the data are drawn from.
Therefore, if we let $\{\bm{x}_i \in S_i \}_{i=1}^{N}, \{(\bm{x}_{n_i}, p_i) \in S_i \times \{1, ..., P\} \}_{i=1}^{N_{sum}}$ denote the optimal data queried from safe active learning criterion eq.~\ref{prob-safe_al}, then the followings still hold:
\begin{align*}
I \left( \bm{Y}_{\phi}, \{ f_{p_k}(\bm{x}_{n_k}) \}_{k=1}^{N_{sum}} \right) &=
\frac12 \sum_{k=1}^{N_{sum}} log \left( 1 + \frac{1}{\sigma_{p_k}^{2}} \Sigma_{k-1}(\bm{x}_{n_k}, p_k) \right), \\
I \left( \bm{Y}, \{ \bm{f}(\bm{x}_{k}) \}_{k=1}^{N} \right) &= \frac12 \sum_{k=1}^{N} log \left( \big| I_P + diag(\{\sigma_i^{2}\}_{i=1}^{P})^{-1} \Sigma_{k-1}(\bm{x}_{k}) \big| \right).
\end{align*}

\subsection{Proof of theorem \ref{thm-convergence_guarantee_safe} - step 2: bound of predictive uncertainty still holds}

We note that $S_i \subseteq \mathcal{X}$ is the safe regions at iteration $i$ determined by the safe model, $\hat{\bm{x}}_i, \bm{x}_i \in S_i$, and $(\hat{\bm{x}}_{n_i}, \hat{p}_i), (\bm{x}_{n_i}, p_i) \in S_i \times \{1, ..., P\}$. We further know from our safe query criterion that

\begin{align*}
\Sigma_{k-1}(\hat{\bm{x}}_{n_k}, \hat{p}_k) &\leq \Sigma_{k-1}(\bm{x}_{n_k}, p_k),\\ |\Sigma_{n-1}(\hat{\bm{x}}_{n})| &\leq |\Sigma_{n-1}(\bm{x}_{n})|.
\end{align*}

Then following the same procedure as Proof of lemma \ref{lemma-predictive_cov_bound}, we obtained the same inequality
\begin{align*}
\frac{1}{N_{sum}}\sum_{k=1}^{N_{sum}}\Sigma_{k-1}(\hat{\bm{x}}_{n_k}, \hat{p}_k)
&\leq \mathcal{O} \left( \frac{1}{N_{sum}} I \left( \bm{Y}_{\phi}, \{ f_{p_k}(\bm{x}_{n_k}) \}_{k=1}^{N_{sum}} \right) \right) ,\\
\frac{1}{N}\sum_{n=1}^{N}|\Sigma_{n-1}(\hat{\bm{x}}_{n})| 
&\leq \mathcal{O} \left( \frac{1}{N} I \left( \bm{Y}, \{ \bm{f}(\bm{x}_{n}) \}_{n=1}^{N} \right) \right).
\end{align*}

Keep in mind that the GP prior is defined from the original data space $\mathcal{X}$.

\subsection{Proof of theorem \ref{thm-convergence_guarantee_safe} - step 3: bound mutual information and maximum information gain}

The same as how we just proved theorem \ref{thm-convergence_guarantee} (section~\ref{supp_section-proof_of_thm2}), we apply Fischer's inequality and theorems in \cite{Srinivas_2012}, and then we obtain the convergence guarantee again.

Notice that $\gamma_p^{N_p} =  max_{\mathcal{D} \subseteq \mathcal{Y}_p(\mathcal{X}): |\mathcal{D}|=N_p} I(\mathcal{D}, f_p)$ and $S_i \subseteq \mathcal{X}$ for each $i$.

\section{Extended Theoretical Result}\label{supp_section-extended_thm_convolution_process}

As the second multi-output model, we consider the convolution processes~\citep{higdon2002space, JMLR:v12:alvarez11a}. 
With the same latent GPs, $\bm{g}: \mathbb{R}^D \rightarrow \mathbb{R}^L$(see \ref{section-GP}), and additionally mappings, $G: \mathbb{R}^D \rightarrow \mathbb{R}^{P \times L}$, that act as a smoothing kernel.
The model becomes
\begin{align*}
\bm{f}(\bm{x}) = \int G(\bm{x}-\bm{z}) \bm{g}(\bm{z}) d\bm{z}.
\end{align*}
The covariance $cov(f_{p}(\bm{x}), f_{p'}(\bm{x}'))$ here is 
%an integral (shown in eq.~\ref{supp_eqn-convolution_process_cov}).
\begin{align*}
\sum_{l=1}^{L}\int \int G_{p,l}(\bm{x}-\bm{z}) G_{p',l}(\bm{x}'-\bm{z}')k_l(\bm{x}, \bm{x}') d\bm{z}d\bm{z}'.
\end{align*}
The smoothing kernel $G$ is usually selected such that this integral in the covariance function is analytically tractable. 
We show that the convergence guarantee we previously got in section~\ref{section-thms} also exists for a convolution process.

\begin{theorem}\label{thm-convolution_process_sal_bound}
	We use $n$ as the unified expression of $N$ and $N_{sum}$. Let $\{\bm{\hat{x}_i} \in S_i \}_{i=1}^{n}$ be $n$ arbitrary inputs drawn from iteration-dependent safe regions $S_i \subseteq \mathcal{X}$, in a partial output setting let $\{\hat{p}_i\}_{i=1}^{n}$ be $n$ arbitrary output component indices. 
	Let $\{\Sigma_{k-1}(\hat{\bm{x}}_k), \Sigma_{k-1}(\hat{\bm{x}}_k, \hat{p}_k)\}$ be the predictive (co)variance of $\bm{\hat{\bm{x}}_k}$ conditioning on $k-1$ training data queried with maximal determinant or entropy under safety constraint (eq. \ref{prob-safe_al}).
	
	If $\int \int |G_{p,l}(\bm{x}-\bm{z}) G_{p,l}(\bm{x'}-\bm{z'})| d\bm{z'} d\bm{z}$ and $k_l(\cdot, \cdot)$ are bounded for all $p$ and $l$, hyperparameters $\bm{\theta}$ are fixed, and $cov(f_{p}(\cdot), f_{p'}(\cdot)) \leq 1$, then 
	\begin{equation*}
	\frac{1}{n}\sum_{k=1}^{n}|\Sigma_{k-1}(\bm{\hat{x}_k})|, \ \frac{1}{n}\sum_{k=1}^{n} \Sigma_{k-1}(\bm{\hat{x}_k}, \hat{p}_k) \leq \mathcal{O} \left( \frac{1}{n} \sum_{p=1}^{P} \gamma_p^{N_p} \right),
	\end{equation*}
	where $\gamma_p^{N_p} =  max_{\mathcal{D} \subseteq \mathcal{Y}_p(\mathcal{X}): |\mathcal{D}|=N_p} I(\mathcal{D}, f_p)$ is the maximum information gain of the current GP $f_p$ on $\mathcal{X}$.
	
	If we furthermore assume $G_{p,l}(\bm{z}) = W_{p,l} \mathcal{N} (\bm{z} \vert \bm{0}, A_p^{-1})$ and $k_l(\bm{z}, \bm{z}') = c_l\mathcal{N} (\bm{z} - \bm{z}' \vert \bm{0}, \Lambda_l^{-1})$ where $ A_p$ and $\Lambda_l$ are positive definite Hermitian matrices, then
	$%\begin{equation*}
	\frac{1}{n} \sum_{p=1}^{P} \gamma_p^{N_p} \leq \mathcal{O} \left( \frac{(log n)^{D+1}}{n} \right).
	$%\end{equation*}
\end{theorem}

Given Gaussian smoothing kernels $G$ and Gaussian latent kernels,
a closed-form expression of the MO covariance function is provided in~\cite{JMLR:v12:alvarez11a} (see also eq.~\eqref{supp_eqn-convolution_process_normal_cov}).
%~\cite{JMLR:v12:alvarez11a} provided %change back if we need the space
%\begin{align*}
%&cov(f_p(\bm{x}), f_{p'}(\bm{x}')) =
%\\&\sum_{l=1}^L W_{p, l} W_{p',l} c_l \mathcal{N} \left( \bm{x} - %\bm{x}' \vert A_p^{-1} + A_{p'}^{-1} + \Lambda_l^{-1} \right).
%\end{align*}
%The corresponding kernel is equivalent to the SE kernel with multivariate lengthscale matrix.
The idea of the proof is identical as for the LMC with only minor differences that we detail out in section~\ref{supp_section-proof_of_thm4}.
%The proof of the first part follows lemma~\ref{lemma-predictive_cov_bound} and first half of theorem~\ref{thm-convergence_guarantee_safe}, and the remaining is similar to the result of theorem~\ref{thm-convergence_guarantee} with minor difference due to the fact that the lengthscale matrices of individual compositions depend on $p$, $p'$ and $l$.

\section{Proof of theorem \ref{thm-convolution_process_sal_bound}}\label{supp_section-proof_of_thm4}

\subsection{Proof of theorem \ref{thm-convolution_process_sal_bound} - bound of uncertainty}

Let $\hat{w}_{p,l}$ denote the bounds of $\int \int |G_{p,l}(\bm{x}-\bm{z}) G_{p,l}(\bm{x'}-\bm{z'})| d\bm{z'} d\bm{z} $ for all $p$ and $l$, and let $\hat{w} = max_{p,l}\{\hat{w}_{p,l}\}$.
Let $\hat{v}$ denote the bound of $k_l(\bm{x}, \bm{x'})$, i.e. $0 \leq k_l(\bm{x}, \bm{x'}) \leq \hat{v}$ for all $l$.
Notice that $|k_l(\bm{x}, \bm{x'})| = k_l(\bm{x}, \bm{x'})$.

\begin{align}
cov(f_p(\bm{x}), f_{p}(\bm{x}'))
&= \sum_{l=1}^{L}\int \int G_{p,l}(\bm{x}-\bm{z}) G_{p',l}(\bm{x}'-\bm{z}')k_l(\bm{x}, \bm{x}') d\bm{z}d\bm{z}' \label{supp_eqn-convolution_process_cov}
\\&\leq \sum_{l=1}^{L} \int \int |G_{p,l}(\bm{x}-\bm{z}) G_{p,l}(\bm{x'}-\bm{z'})| |k_l(\bm{x}, \bm{x'})| d\bm{z'} d\bm{z} \nonumber
\\&\leq \sum_{l=1}^{L} \hat{w} k_l(\bm{x}, \bm{x'}) \nonumber
\\&\leq \sum_{l=1}^{L} \hat{w} \hat{v} \nonumber
\\&= L \hat{w} \hat{v}. \nonumber
\end{align}

Now we can follow the proof of lemma \ref{lemma-predictive_cov_bound} (section~\ref{supp_section-proof_of_main_lemma1}).
Notice here that lemma \ref{supp_lemma-mutual_information}, which is used in section~\ref{supp_section-proof_of_main_lemma1} to obtain the mutual information term, is independent of kernel function.
Set 
\begin{align*}
(a, b) &=\left(\frac{1}{\sigma_{p_k}}\Sigma_{k-1}(\cdot, \cdot), \frac{L \hat{w} \hat{v}}{\sigma_{p_k}}\right) \text{ or } \left(\frac{1}{\prod_{p=1}^{P} \sigma_p^2}|\Sigma_{k-1}(\cdot)|, \frac{ L \hat{w} \hat{v}}{\prod_{p=1}^{P} \sigma_p^2}\right),
\end{align*}
then $\frac{a}{\log (1+a)} \leq \frac{b}{\log (1+b)}$, so eq.~\eqref{supp_eqn-predictive_uncertainty_mutual_info_ineq_poo}~\eqref{supp_eqn-predictive_uncertainty_mutual_info_ineq_constant}~\eqref{supp_eqn-predictive_uncertainty_mutual_info_ineq} give us
\begin{align*}
\frac{1}{N_{sum}}\sum_{k=1}^{N_{sum}}\Sigma_{k-1}(\cdot, \cdot)
&\leq \mathcal{O} \left( \frac{1}{N_{sum}} I \left( \bm{Y}_{\phi}, \{ f_{p_k}(\bm{x}_{n_k}) \}_{k=1}^{N_{sum}} \right) \right),
\\&\leq \mathcal{O} \left( \frac{1}{N_{sum}} \sum_{p=1}^{P} \gamma_p^{N_p} \right),\\
\frac{1}{N}\sum_{n=1}^{N}|\Sigma_{n-1}(\cdot)|
&\leq \mathcal{O} \left( \frac{1}{N} I \left( \bm{Y}, \{ \bm{f}(\bm{x}_{n}) \}_{n=1}^{N} \right) \right)
\\&\leq \mathcal{O} \left( \frac{1}{N} \sum_{p=1}^{P} \gamma_p^{N_p} \right) \text{, here } N_p = N.
\end{align*}

\subsection{Proof of theorem \ref{thm-convolution_process_sal_bound} - bound $\gamma_p^{N_p}$ for the given kernel}

If $G_{p,l}(\bm{z}) = W_{p,l} \mathcal{N} (\bm{z} \vert \bm{0}, A_p^{-1})$ and $k_l(\bm{z}, \bm{z}') \propto \mathcal{N} (\bm{z} - \bm{z}' \vert \bm{0}, \Lambda_l^{-1})$, then from~\cite{JMLR:v12:alvarez11a} we have
\begin{align}\label{supp_eqn-convolution_process_normal_cov}
cov(f_p(\bm{x}), f_{p'}(\bm{x}'))
&=\sum_{l=1}^L W_{p, l} W_{p',l} c_l \mathcal{N} \left( \bm{x} - \bm{x}' \vert A_p^{-1} + A_{p'}^{-1} + \Lambda_l^{-1} \right)
\\&\propto \sum_{l=1}^L c_{p, p',l} \mathcal{N} \left( \bm{x} - \bm{x}' \vert \bm{0}, A_p^{-1} + A_{p'}^{-1} + \Lambda_l^{-1} \right),
\end{align}
where each of $c_{p, p',l}$ is a scalar parameter.
Here the kernel is a sum of latent kernels dependent not only of $l$ but also of $p$ (and thus this model provides more flexibility than LMC).

For each $p$, we are actually dealing with $\gamma_p^{N_p} =  max_{\mathcal{D} \subseteq \mathcal{Y}_p(\mathcal{X}): |\mathcal{D}|=N_p} I(\mathcal{D}, f_p)$ where $f_p$ is a GP with kernel $cov(f_p(\bm{x}), f_p(\bm{x}'))$, which is a weighted sum of SE kernels with matrix lengthscales. This can be seen by normalizing $\mathcal{N} \left( \bm{x} - \bm{x}' \vert A_p^{-1} + A_{p'}^{-1} + \Lambda_l^{-1} \right)$, which gives us 
\begin{align*}
cov(f_p(\bm{x}), f_{p}(\bm{x}')) \propto \sum_{l=1}^L \tilde{c}_{p, p,l} exp \left( (\bm{x} - \bm{x}')^T (A_p^{-1} + A_p^{-1} + \Lambda_l^{-1})^{-1} (\bm{x} - \bm{x}') \right).
\end{align*}

Therefore, we follow the previous proof for SE kernels:
\begin{enumerate}
	\item For each $p$ and $l$, section~\ref{supp_section-bound_for_SE_Matrix_lengthscale} bounds the eigenvalues of $exp \left( (\bm{x} - \bm{x}')^T (A_p^{-1} + A_p^{-1} + \Lambda_l^{-1})^{-1} (\bm{x} - \bm{x}') \right)$.
	\item Follow section~\ref{supp_section-bound_for_SE} to get $\gamma_p^{N_p} \leq \mathcal{O}((\log N_{sum})^{D+1})$ (eq.~\eqref{supp_eqn-gamma_bound_for_SE}) for the GP  $f_p$.
	\item As this bound does not depend on $p$, we again obtain $\frac{1}{N_{sum}}\sum_{p=1}^{P}\gamma_p^{N_p} \leq \mathcal{O}\left( \frac{(\log N_{sum})^{D+1}}{N_{sum}} \right)$.
\end{enumerate}

\section{Experimental Details}\label{supp_section-exp_detail}

\begin{figure}[h]
	\vspace{.3in}
	\centerline{
		\includegraphics[width=0.32\textwidth]{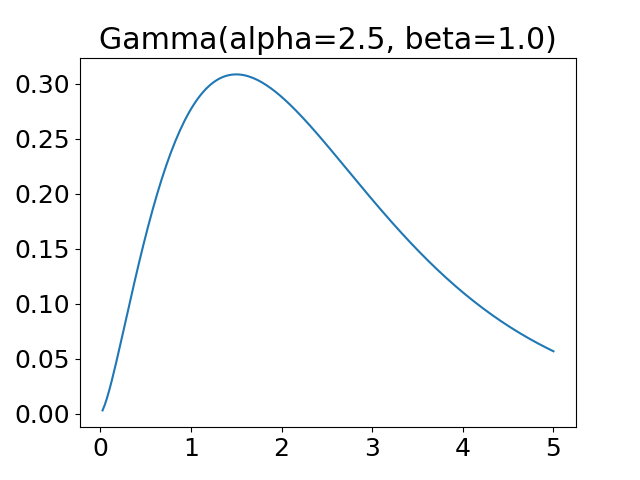}
		\includegraphics[width=0.32\textwidth]{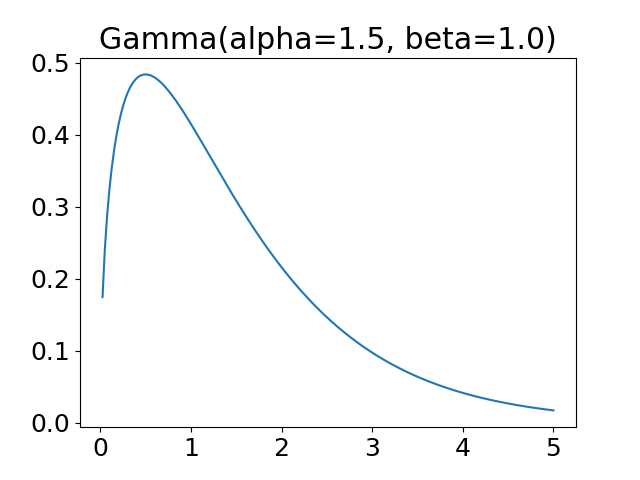}
		\includegraphics[width=0.32\textwidth]{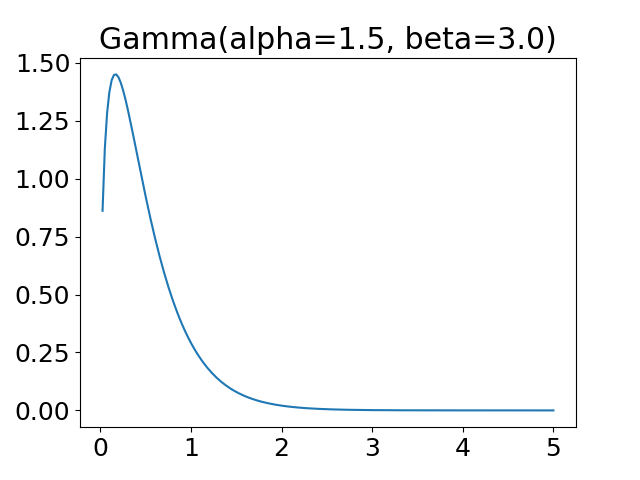}
	}
	\centerline{
		\includegraphics[width=0.3\textwidth]{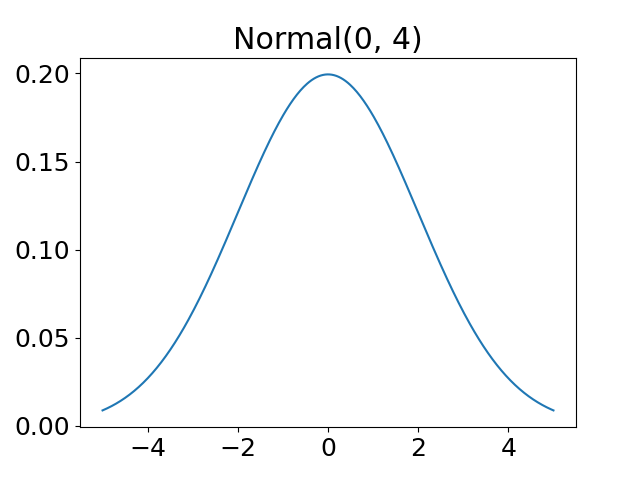}
	}
	\caption{
		Probability density functions for Bayesian treatment. The distributions are priors of variance (top left) and lengthscale (top middle) of latent kernels, observation noise variances (top right) and kernel weight $W_{pl}$ (bottom). X-axis is the value of each random variable and y-axis is the probability density.
	}
	\label{supp_fig-priors}
\end{figure}

\begin{figure}[h]
	\vspace{.3in}
	\centerline{\fbox{
			\includegraphics[width=0.95\columnwidth]{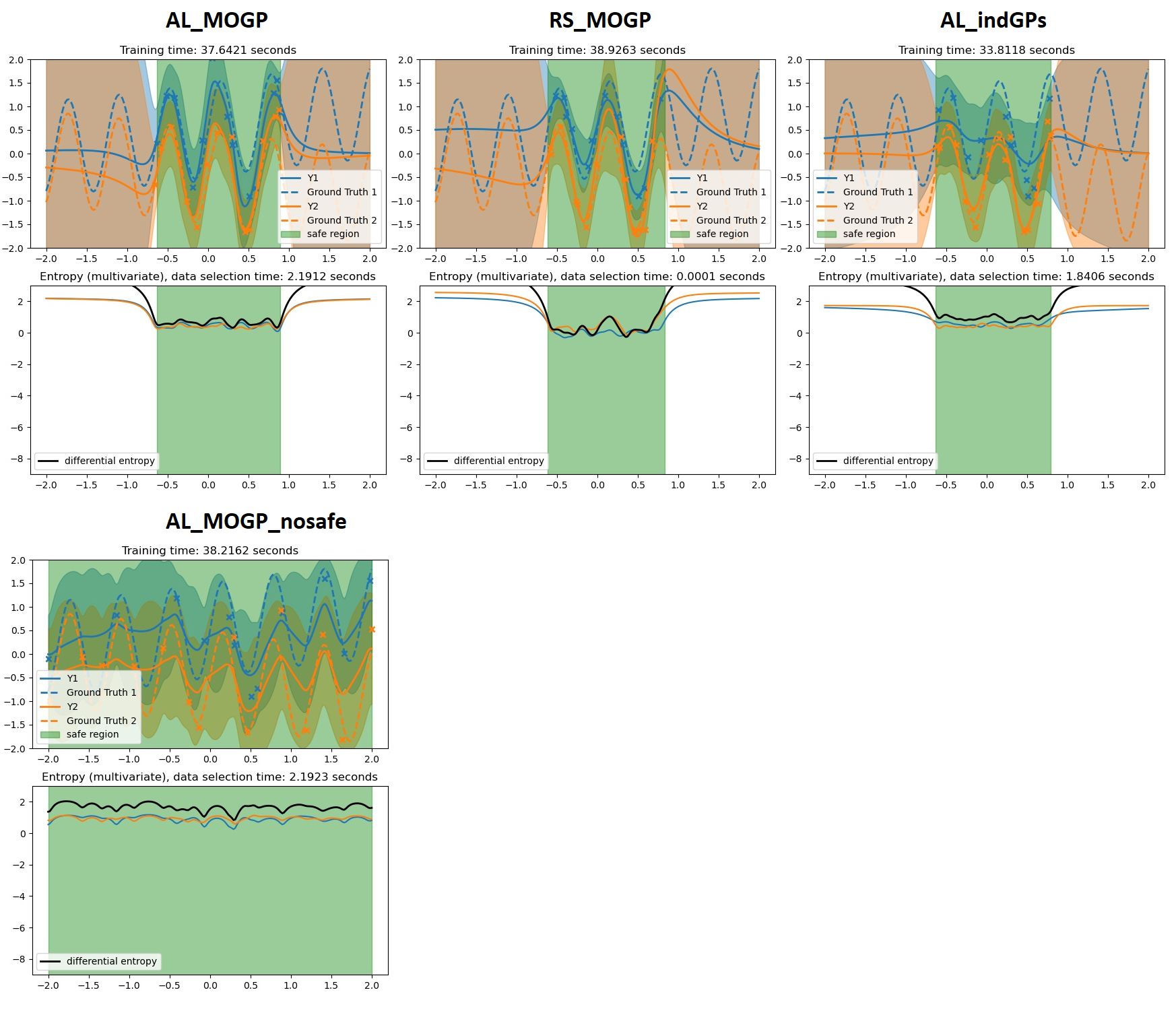}
	}}
	\vspace{.3in}
	\caption{
		Different pipelines in sin \& sigmoid simulation at the 15-th iteration. The plots demonstrate model predictions, ground truth and observed data (first row) as well as entropy (second row) of different frameworks. AL\_MOGP\_nosafe is achieved by setting the safety threshold such that it is safe everywhere. The colors indicate the 2 outputs. Entropy of full output covariance is shown with black lines. Training time here is the sampling time of Bayesian treatment (HMC).
	}
	\label{supp_fig-toyset}
	
	\vspace{.3in}
	\centerline{\fbox{
			\includegraphics[width=0.95\columnwidth]{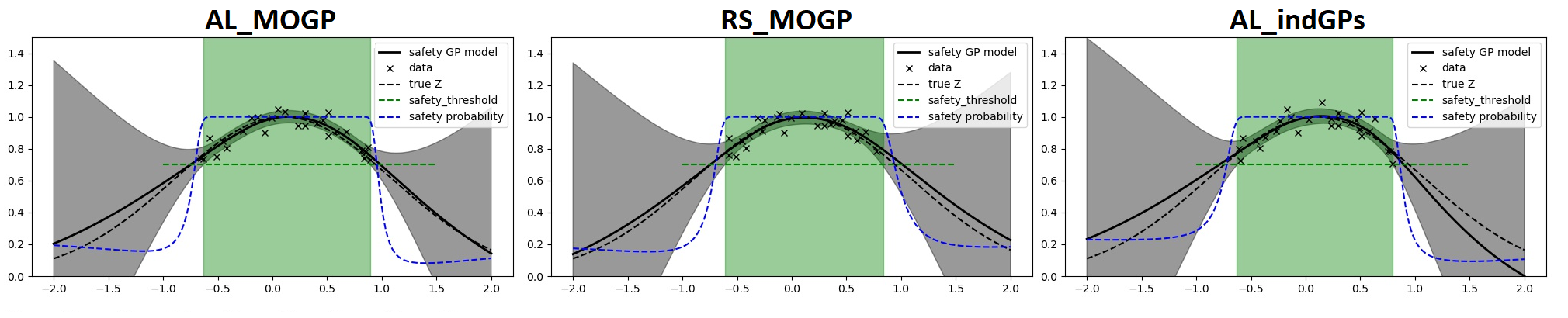}
	}}
	\vspace{.3in}
	\caption{
		Safety control for different pipeline in simulation 1 at the 15-th iteration. The plots are model predictions, ground truth safety values, observed safety data and the safe probability given by the GP model.
	}
	\label{supp_fig-toyset_safety}
\end{figure}

\begin{figure}[h]
	\vspace{.3in}
	\centerline{\fbox{
			\includegraphics[width=0.95\columnwidth]{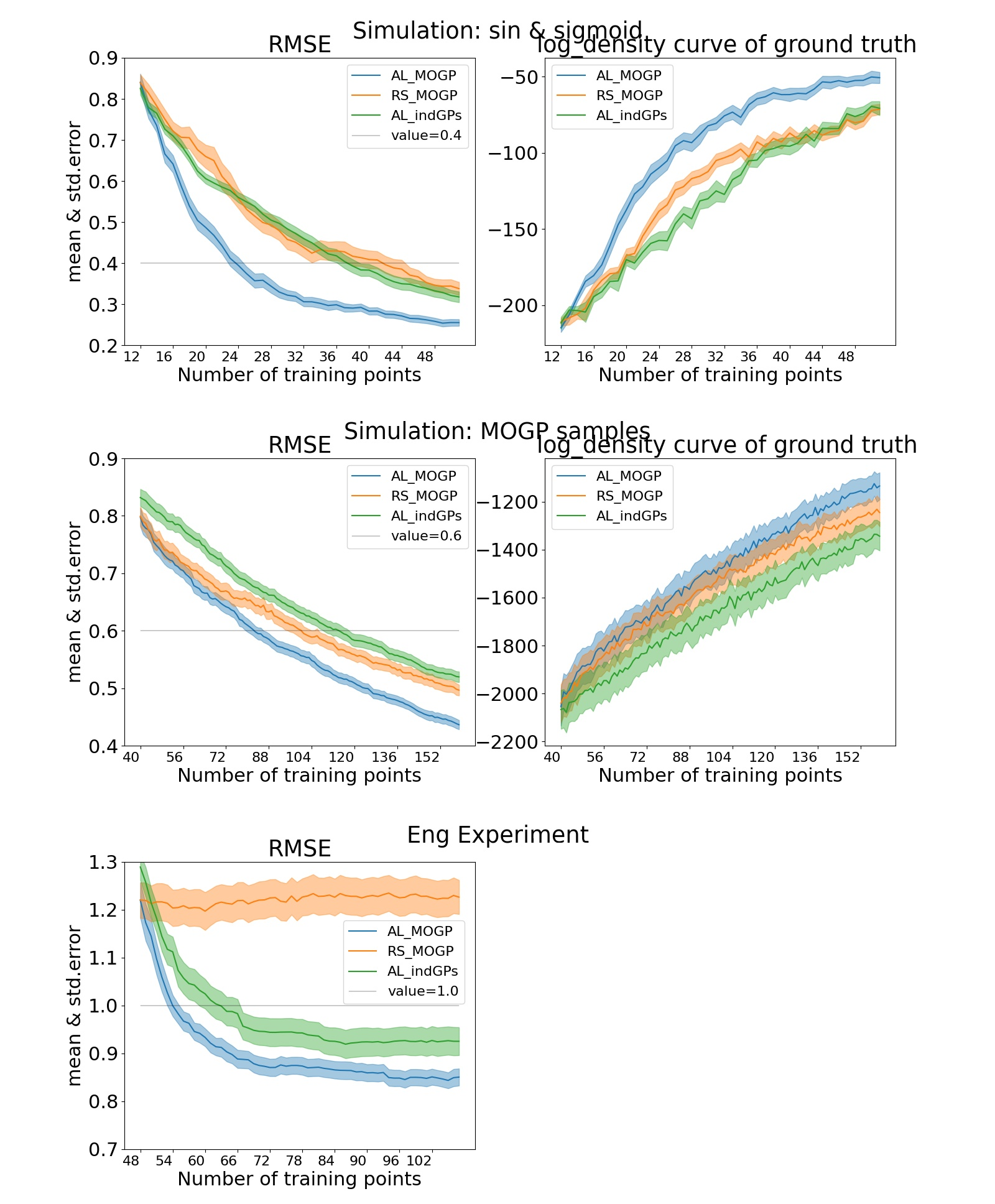}
	}}
	\vspace{.3in}
	\caption{
		RMSE and log density of test data. The left column is the same as figure~\ref{fig1}. In sin \& sigmoid simulation experiment, the test data are ground truth values. The test data used for evaluation are all safe data. 
	}
	\label{supp_fig-RMSE_log_dens}
\end{figure}

\begin{figure}[h]
	\vspace{.3in}
	\centerline{\fbox{
			\includegraphics[width=0.48\textwidth]{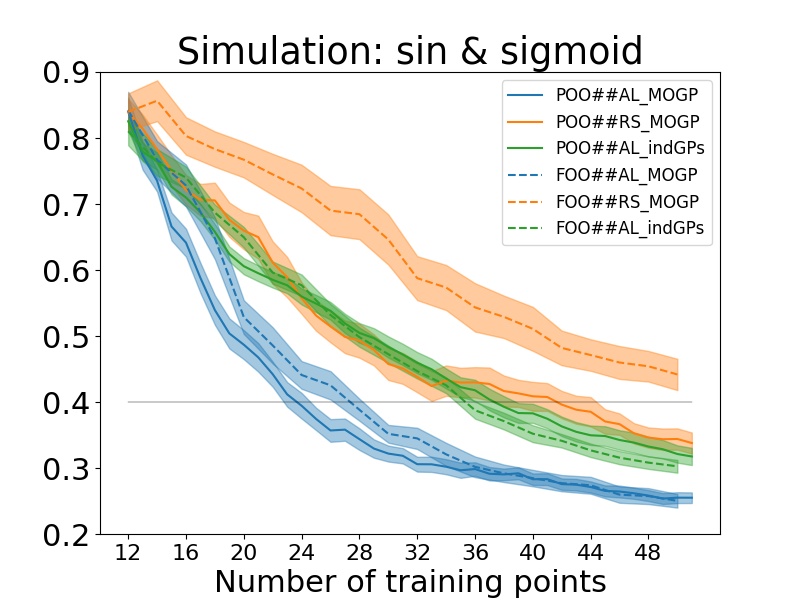}
			\includegraphics[width=0.48\textwidth]{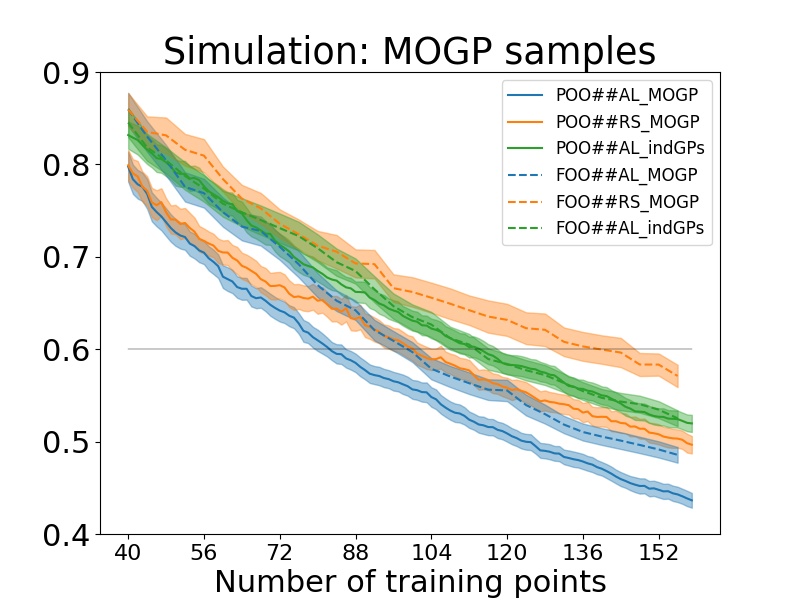}
		}
	}
	
	\caption{
		RMSE of different pipelines with partially observed outputs (POO) and with fully observed outputs (FOO). 
		POO curves are identical to those shown in figure~\ref{fig1}. Y-axis is the RMSE value and x-axis is $N_{sum}$.
	}
	\label{supp_fig-POOvsFOO_sims}
\end{figure}

%\begin{figure}[h]
%	\vspace{.3in}
%	\centerline{\fbox{
%			\includegraphics[width=0.48\textwidth]{figures/Scom_3_OWEOdata_HC_O2_RMSE_mean_safe_global_POOvsFOO.png}			
%			\includegraphics[width=0.48\textwidth]{figures/Scom_3_OWEOdata_HC_O2_POO_queried_task.png}
%		}
%	}
%	
%	\caption{
%		Left: RMSE of different pipelines with partially observed outputs (POO) and with fully observed outputs (FOO). POO curves are identical to those shown in figure~\ref{fig2}. Y-axis is the RMSE value and x-axis is $N_{sum}$.\\
%		Right: Portions of the second output $\{y_{2,n} | n =1, ...\}$ among all query $\{y_{p,n} | p=1 or 2, n=1, ...\}$ over one experiment for AL\_MOGP. The histogram is the distribution of such portions over different seeds. Y-axis is the count of seeds and x-axis is the portion. This plot shows that most of the queries are for the first output (HC).
%	}
%	\label{supp_fig-POOvsFOO_OWEO}
%\end{figure}

\begin{table}[b]
	\parbox[c]{\linewidth}{\centering
	\caption{Toy dataset, safety model precisions} \label{table-safety_toy}
	\begin{tabular}{r|ccc}
		\textbf{$N_{sum}$} &\textbf{AL\_MOGP} &\textbf{RS\_MOGP} &\textbf{AL\_indGPs}\\
		\hline \\
		12  & $0.9925 \pm 0.0027$ & $0.9930 \pm 0.0019$ & $ 0.9943 \pm 0.0018$\\
		22 & $0.9895 \pm 0.0027$ & $0.9930 \pm 0.0019$ & $ 0.9917 \pm 0.0020$\\
		32 & $0.9913 \pm 0.0021$ & $0.9906 \pm 0.0024$ & $ 0.9901 \pm 0.0022$\\
		42 & $0.9901 \pm 0.0021$ & $0.9921 \pm 0.0022$ & $ 0.9893 \pm 0.0023$\\
	\end{tabular}
	}
	%\hspace{0.05\linewidth}
	\bigskip
	
	\parbox[c]{\linewidth}{\centering
	\caption{GP dataset, safety model precisions} \label{table-safety_GP}
	\begin{tabular}{r|ccc}
		\textbf{$N_{sum}$} &\textbf{AL\_MOGP} &\textbf{RS\_MOGP} &\textbf{AL\_indGPs}\\
		\hline \\
		40   & $0.9980 \pm 0.0006$ & $0.9980 \pm 0.0006$ & $0.9979 \pm 0.0007$\\
		60  & $0.9979 \pm 0.0009$ & $0.9978 \pm 0.0005$ & $0.9987 \pm 0.0003$\\
		80  & $0.9983 \pm 0.0003$ & $0.9981 \pm 0.0004$ & $0.9985 \pm 0.0003$\\
		100 & $0.9981 \pm 0.0004$ & $0.9976 \pm 0.0006$ & $0.9981 \pm 0.0004$\\
		120 & $0.9975 \pm 0.0005$ & $0.9976 \pm 0.0004$ & $0.9978 \pm 0.0005$\\
		140 & $0.9977 \pm 0.0005$ & $0.9975 \pm 0.0004$ & $0.9976 \pm 0.0005$\\
	\end{tabular}
	}
	%\hspace{0.05\linewidth}
	\bigskip
	
	\parbox[c]{\linewidth}{\centering
	\caption{EngE dataset, safety model precisions} \label{table-safety_OWEO}
	\begin{tabular}{r|ccc}
		\textbf{$N_{sum}$} &\textbf{AL\_MOGP} &\textbf{RS\_MOGP} &\textbf{AL\_indGPs}\\
		\hline \\
		48  & $0.9841 \pm 0.0020$ & $0.9841 \pm 0.0020$ & $0.9843 \pm 0.0020$\\
		58 & $0.9828 \pm 0.0019$ & $0.9848 \pm 0.0022$ & $0.9827 \pm 0.0020$\\
		68 & $0.9829 \pm 0.0019$ & $0.9869 \pm 0.0017$ & $0.9835 \pm 0.0017$\\
		78 & $0.9835 \pm 0.0017$ & $0.9876 \pm 0.0014$ & $0.9831 \pm 0.0018$\\
		88 & $0.9830 \pm 0.0018$ & $0.9880 \pm 0.0013$ & $0.9832 \pm 0.0018$\\
		98 & $0.9841 \pm 0.0017$ & $0.9890 \pm 0.0010$ & $0.9839 \pm 0.0018$\\
	\end{tabular}
	}
	
	%\begin{center}
	\caption*{
		We define the precision of the safety model as the fraction of samples that are within the true safe region from all samples that are marked as probabilistically safe from the safety model.
		The tables report the mean and standard error over this statistic for each experimental pipeline (AL\_MOGP, RS\_MOGP, AL\_indGPs) and for each dataset.
		On both datasets, all methods fulfill the safety criterion after the first iteration.
		%A precision ratios of true positive to all positive predictions, where positive is safe for us.
		%The safety models operate on the same input domain $\bm{X}$ as the corresponding main models, but the inference is only on the safety values $\bm{Z}$ (see section~\ref{section-safety_AL}). 
		%See figure~\ref{fig1} for the performance of the main models on $\bm{Y}_{\phi}$. 
		Table~\ref{table-safety_toy} corresponds to the toy dataset and table~\ref{table-safety_OWEO} to the EngE dataset.
	}
	%\end{center}
\end{table}

In each experiment, we randomly select a number of data as a initial dataset.
With this initial dataset, we run algorithm~\ref{alg-SAL} for AL\_MOGP, AL\_indGPs, RS\_MOGP and the no-safety reference AL\_MOGP\_nosafe.
Therefore, in each experiment, the initial dataset is always the same for all frameworks.
Notice that for RS\_MOGP, we query a random point under safety constraint.
AL\_indGPs is equivalent to our AL\_MOGP with $L=P$ and $W=I_P$, see section~\ref{section-GP}.

For all of our models, we use M{\'a}tern kernels with $\nu=\frac{5}{2}$~\citep{3569} for $k_l$.
In this paper, the models always use $L$ equal to $P$.

\subsection{Inference with Hyperparameters}\label{supp_section-exp_detail_hyperparameters_tuning_detail}

The hyperparameters, i.e. kernel variances, kernel lengthscales and observation noise variance(s), are denoted jointly by $\bm{\theta}$.

\paragraph{Type II maximum likelihood estimation}

The log marginal likelihoods are 
\begin{equation}\label{supp_obj-log_marginal_likelihood}
\mathcal{L}(\bm{\theta}, \mathcal{D}) %= log p(Y | \bm{X}, \bm{\theta})
= log \mathcal{N}(Y | \bm{0}, K_{NN} + \sigma^2 I_N) \text{ and }
\end{equation} 
\begin{equation}\label{supp_obj-mo_log_marginal_likelihood}
\mathcal{L}(\bm{\theta}, \mathcal{D}) %= log p(\bm{Y} | \bm{X}, \bm{\theta})
= \log \mathcal{N} \left(\bm{Y}_{\phi} \big| \bm{0}, \Omega_{N_{sum}N_{sum}} + \diag(\{\sigma_{p_k}^2\}_{k=1}^{N_{sum}}) \right)
\end{equation} 
for GP regression and for MOGP regression, respectively. 
Here $K_{NN} + \sigma^2 I_N$ and $\Omega_{N_{sum}N_{sum}} + \diag(\{\sigma_{p_k}^2\}_{k=1}^{N_{sum}})$ are functions of the hyperparameters $\bm{\theta}$ (see also eq.~\eqref{eqn-gp_mean}-\eqref{eqn-gp_variance}, eq.~\eqref{eqn-mo_posterior_mu}-\eqref{eqn-mo_posterior_cov_poo}).
Computing the likelihoods have complexities $\mathcal{O}(N^3)$ and $\mathcal{O}(N_{sum}^3)$ as the inversion of the covariance matrices is required.

The hyperparameters
\begin{align*}
\hat{\theta} = argmax_{\bm{\theta}} \mathcal{L}(\bm{\theta}, \mathcal{D})
\end{align*}
can be obtained by applying gradient based methods.
Then the predictions can be simply done by substituting $\hat{\theta}$ into the models and apply eq.\eqref{eqn-gp_mean}-\eqref{eqn-gp_variance} for standard GP regression and with eq.~\eqref{eqn-mo_posterior_mu}-\eqref{eqn-mo_posterior_cov_poo} for MOGP regression.

\paragraph{Bayesian treatment-theoretics}\label{supp_section-exp_detail_bayesian_treatment} \ 

To perform a Bayesian treatment, we first assign prior distributions over the hyperparameters, i.e. $p(\bm{\theta})$, apply Bayes rule to $p(\bm{\theta})$ and $\mathcal{L}(\bm{\theta}, \mathcal{D})$ to obtain $p \left( \bm{\theta} \vert \mathcal{D} \right)$, and then the prediction becomes:
\begin{equation}\label{supp_eqn-true_posterior}
p(\bm{f}(\bm{x}_{*}) \vert \bm{x}_{*}, \mathcal{D})  
=
\int p \left( \bm{f}(\bm{x}_{*}) | \bm{x}_{*}, \mathcal{D}, \bm{\theta} \right) p \left( \bm{\theta} | \mathcal{D} \right) d\bm{\theta}.
\end{equation}
Here $p \left( \bm{f}(\bm{x}_{*}) | \bm{x}_{*}, \mathcal{D}, \bm{\theta} \right)$ is the GP posterior given hyperparameters (see eq.~\eqref{eqn-gp_mean}-\eqref{eqn-gp_variance}, eq.~\eqref{eqn-mo_posterior_mu}-\eqref{eqn-mo_posterior_cov_poo}).
Notice that the integral is intractable, and we either need to perform approximate inference~\citep{titsias2014doubly} or resort to Monte Carlo sampling. 
In our work, we apply the latter and approximate eq.~\eqref{supp_eqn-true_posterior} by drawing samples from the posterior
\begin{align}\label{supp_eqn-HMC_posterior}
p(\bm{f}(\bm{x}_{*}) \vert \bm{x}_{*}, \mathcal{D})
&\approx \frac{1}{|\{\hat{\theta}\}|} \sum_{\hat{\theta}} p \left( \bm{f}(\bm{x}_{*}) | \bm{x}_{*}, \mathcal{D}, \hat{\theta} \right),
%\\
%&=\mathcal{N} \left(
%\frac{1}{|\{\hat{\theta}\}|} \sum_{\hat{\theta}} \mu(\bm{x}_*, \hat{\theta}),
%\frac{1}{|\{\hat{\theta}\}|} \sum_{\hat{\theta}} \Sigma(\bm{x}_*, \hat{\theta})
%\right), \nonumber
\end{align}
where $\hat{\theta}$ are drawn from $p(\bm{\theta} | \mathcal{D}) \propto \mathcal{L}(\bm{\theta}, \mathcal{D}) p(\bm{\theta})$. 

Assuming that different hyperparameters are independent, $p(\bm{\theta})$ is the product of priors of individual hyperparameters.
The priors of hyperparameters are Gamma and Normal distributions, which are also shown in figure~\ref{supp_fig-priors}:
\begin{itemize}
	\item $\Gamma(\alpha=2.5, \beta=1.0)$ for latent kernels $k_l(\cdot)$, variance,
	\item $\Gamma(1.5, 1.0)$ for latent kernels $k_l(\cdot)$, lengthscale,
	\item $\Gamma(1.5, 3)$ for noise variances $\sigma_p^2$ and
	\item $\mathcal{N}(0, 2^2)$ for $W_{pl}$.
\end{itemize}

Gamma priors are selected for positive parameters.
$\alpha=1.5$ and $\alpha=2.5$ would push the distribution mean further from $0$.
As all the variances are assumed to be bounded and the datasets are assumed normalized, the distributions should also be not too far away from $0$.
For observation noise variances, we use larger $\beta$ to encourage smaller values.
For the kernel variances, we use larger $\alpha$ ($2.5$) to encourage large uncertainty, which should be generally true without observing data.
Lengthscales of the kernel can be any value greater than $0$, and the effect of different prior setting does not seem very obvious (experiment not shown).
The kernel variances are still weighted by $W_{pl}$ in the model.
Because $W_{pl}$ are bounded and should be symmetric to $0$, we place a normal distribution centering around $0$.

In our initial experiments, we tried few different prior parameters, but the effect did not seem obvious.
We also use the same hyperpriors over all experiments.
For the safety model, a large kernel variance ensure that the probabilistic safety condition is difficult to achieve.
In this case, the model can only have high confidence with enough observations, and this is desired for the safety model.
The prior for safety model could also be set according to the safety threshold.
For example, if it is safe to have safety value greater than $1$, one could consider a prior with mean larger than $1$ or even $2$ for kernel variances, or, in addition, adjust the GP to non-zero mean and the prior for GP mean could be set to encourage values centering around $2$.

\paragraph{Bayesian treatment-implementation} \ 

In this work, we use Hamiltonian Monte Carlo (HMC)~\citep{betancourt2018conceptual, brooks2011handbook} as our sampling method (for approximation eq.~\eqref{supp_eqn-HMC_posterior}). 
We always use 100 hyperparameter vectors for each inference.
We pick 1 sample out of 20 to ensure the samples are sufficiently independent, and we abandon the first 300 samples to ensure all the samples actually lie on the target distribution.
Therefore, for each inference, we sample 100 hyperparameter vectors out of a chain of 2300 samples.

For a GP model, sampling $T_{\theta}$ hyperparameters has complexity $\mathcal{O}(T_{\theta})*\mathcal{O}(\mathcal{L}(\bm{\theta})) = \mathcal{O}(T_{\theta}*N_{sum}^3)$, where $\mathcal{O}(T_{\theta})$ absorbs the sampler's setting into a constant (step of the sampler, acceptance rate etc, see~\cite{brooks2011handbook}).
The datasets we use are however small and HMC has a quite good acceptance rate (roughly 0.7) for our model, so this method is not too slow in practice.

For performing inference with HMC method, as making predictions with different hyperparameter sets (and for different points) can be done in parallel, a Bayesian treatment does not necessarily increase the inference time.

The HMC is implemented with tensorflow\_probability (tfp).
The samples are generated with tfp.mcmc.sample\_chain:
\begin{align*}
\text{samples, \_ =} &\text{tfp.mcmc.sample\_chain(}
\\&\text{num\_results=100, num\_burnin\_steps=300, num\_steps\_between\_results=20,}
\\&\text{current\_state=helper.current\_state,}
\\& \text{kernel=tfp.mcmc.SimpleStepSizeAdaptation(}
\\& \ \ \ \ \text{tfp.mcmc.HamiltonianMonteCarlo(}
\\&\ \ \ \ \ \ \ \ \text{target\_log\_prob\_fn=helper.target\_log\_prob\_fn,}
\\&\ \ \ \ \ \ \ \ \text{num\_leapfrog\_steps=10, step\_size=0.01}
\\&\ \ \ \ \ \ \ \ \text{),}
\\& \ \ \ \ \text{num\_adaptation\_steps=int(0.3*300),}
\\& \ \ \ \ \text{target\_accept\_prob=f64(0.75),}
\\& \ \ \ \ \text{adaptation\_rate=0.1}
\\& \ \ \ \ \text{),}
\\&\text{trace\_fn=lambda \_, pkr: pkr.inner\_results.is\_accepted}
\\&\text{)}
\\\text{helper =} & \text{gpflow.optimizers.SamplingHelper(}
\\&\text{log\_posterior\_density (eq.~\ref{supp_obj-mo_log_marginal_likelihood}),} \\&\text{model.trainable\_parameters}
\\&\text{).}
\end{align*}

\subsection{Entropy computation for HMC}\label{supp_section-exp_detail_entropy_approximation} \ 

Given a random variable $y$, entropy
\begin{align*}
H(y) = - \int p(y) \log p(y) dy.
\end{align*}

With the HMC approximation for Bayesian treatment (eq.~\eqref{supp_eqn-HMC_posterior}), the entropy shown as follows is intractable, where $p$ in brackets is the output index while $p$ out of brackets is probability
\begin{align*}
H(\bm{f}(\bm{x}_*), p) = - \int \frac{1}{|\{\hat{\theta}\}|} \sum_{\hat{\theta}} p \left( \bm{f}(\bm{x}_{*}), p | \bm{x}_{*}, \mathcal{D}, \hat{\theta} \right)
\log \left(\frac{1}{|\{\hat{\theta}\}|} \sum_{\hat{\theta}} p \left( \bm{f}(\bm{x}_{*}), p | \bm{x}_{*}, \mathcal{D}, \hat{\theta} \right)\right) d\bm{f}(\bm{x}_*).
\end{align*}

To estimate the entropy efficiently, we use a Gaussian mixture approximation
\begin{align*}
\frac{1}{|\{\hat{\theta}\}|} \sum_{\hat{\theta}} p_* \left( \bm{f}(\bm{x}_{*}) | \bm{x}_{*}, \mathcal{D}, \hat{\theta} \right)
&\approx
\mathcal{N}(\bm{f}(\bm{x}_{*}) | \mu_{HMC}(\bm{x}_{*}, p_*), \Sigma_{HMC} (\bm{x}_{*}, p_*) ),
\\ \mu_{HMC}(\bm{x}_{*}, p_*) &= \frac{1}{|\{\hat{\theta}\}|} \sum_{\hat{\theta}} \mu_{\hat{\theta}}(\bm{x}_{*}, p_*)
\\ \Sigma_{HMC}(\bm{x}_{*}, p_*) &= \frac{1}{|\{\hat{\theta}\}|} \sum_{\hat{\theta}} \left( \Sigma_{\hat{\theta}}(\bm{x}_{*}, p_*) + \mu_{\hat{\theta}}(\bm{x}_{*}, p_*) \mu_{\hat{\theta}}(\bm{x}_{*}, p_*)^T \right) - \mu_{HMC}(\bm{x}_{*}, p_*) \mu_{HMC}(\bm{x}_{*}, p_*)^T,
\end{align*}
which then results in a tractable entropy $\propto \log (|\Sigma_{HMC}(\bm{x}_{*}, p_*)|)$.

\subsection{Experiments with different datasets}

For all of the datasets, we repeat the experiments 30 times and set the safety probability threshold to $\delta=0.05$.
We averaged the RMSE values over the different output components.

\paragraph{Dataset: simulation with sin \& sigmoid}\label{supp_section-exp_detail_toydata} \ 

The data are simulated as follows
\begin{align*}
\bm{f}_{true}(x)&=\begin{pmatrix}
sin(10x) + \frac{1}{1+exp(-2x)}\\
sin(10x) - \frac{1}{1+exp(-2x)}
\end{pmatrix},\\
h_{true}(x)&=exp(-(x-0.1)^2/2),\\
\bm{y} &\sim \mathcal{N}\left( \bm{f}_{true}(x),
\begin{pmatrix}
0.4^2 & 0 \\ 0 & 0.4^2
\end{pmatrix} \right),\\
z &\sim \mathcal{N}\left( \ h_{true}(x), 0.05^2 \right).
\end{align*}
We say $x$ is safe if $h(x) > 0.7$ and set the allowed risk to $\delta=0.05$ (see section~\ref{section-safety_AL}). 
Therefore, in order to be executed, x needs to fullfill $p(h(x) > 0.7) > 0.95$.
The true safety values $h_{true}(x) > 0.7$ are thereby equivalent to the input interval, $x \in \left( 0.1-\sqrt{-2\log (0.7)}, 0.1+\sqrt{-2 \log (0.7)} \right) $ (roughly $(-0.74, 0.94)$).
Interval for exploration is set to $x \in \left[ -2, 2 \right]$.

Figure~\ref{supp_fig-toyset} shows the models, data and entropy of the 3 frameworks at the 15th iteration.
When the output is partially observed which is how the experiments are done, the entropy are the concatenation of $H(\bm{f}(\bm{x}_*), p)$ for all $p$, corresponding to the blue and orange entropy curves in figure~\ref{supp_fig-toyset}.
The corresponding safety predictions are in figure~\ref{supp_fig-toyset_safety}.

In this experiment we start with $N_{sum}=12$ ($6$ for each output).
The RMSE and log likelihood are evaluated on ground truth $\bm{f}_{true}$ of a test set drawn from the safe region.

\paragraph{Dataset: MOGP samples}\label{supp_section-exp_detail_GPdata} \ 

We first fix a seed ($=123$), specify the number of experiments ($E=30$) and the number of data points in each experiment ($N_{training} + N_{test}=2000+500$), and specify the dimension of input ($D=2$), output ($P=4$), and the number of latent GPs ($L=3$).

The goal is to have input $\bm{X} \subseteq \mathbb{R}^D$, output $\bm{Y}_1, ..., \bm{Y}_E \subseteq \mathbb{R}^P$ and safety values $\bm{Z}_1, ..., \bm{Z}_E \subseteq \mathbb{R}$.

This can be done by drawing samples from a given MOGP and GP.
We draw the samples as follows, where the sample interval and kernels can all be replaced, as long as the bounded conditions are fulfilled:
\begin{enumerate}
	\item Input $\bm{X} \subseteq \mathbb{R}^D$: draw $(N_{training} + N_{test}) \times D$ samples uniformly from interval $[-2, 2)$, remove duplicate sample vectors, draw more samples if there were duplicate samples being removed until having $N_{training} + N_{test}$ samples, and then shuffle all vectors to preserve randomness.
	
	\item Prepare kernels for (MO)GPs: draw samples uniformly from interval $[0.01, 1)$ for $L+1$ squared exponential kernels ($k_1(\cdot), ..., k_L(\cdot)$ for samples $\bm{Y}$ and $k_z(\cdot)$ for samples $\bm{Z}$, each with a variance and a scalar lengthscale).
	To normalize the data, we fix the variances of $k_1(\cdot), ..., k_L(\cdot)$ to $1$, and to ensure a smoother safety values we fix the lengthscale of $k_z(\cdot)$ to $1$.
	The safety values does not have to be very smooth, but it is then necessary to analyze how the experiments can start with a robust enough safety model, which is not the focus of this paper (see~\cite{10.1007/978-3-319-23461-8_9} for safety discussion).
	
	\item Draw latent samples and noise-free $\bm{Z}$: draw $E$ $L$-dim trajectories denoted by $\bm{G}_1, ..., \bm{G}_E \subseteq \mathbb{R}^{(N_{training} + N_{test}) \times D}$, individual dimensions following $\mathcal{N}(\bm{0}, k_1(\bm{X}, \bm{X})), ..., \mathcal{N}(\bm{0}, k_L(\bm{X}, \bm{X}))$ , and draw $E$ sets of noise-free $\bm{Z}$ from $\mathcal{N}(\bm{0}, k_z(\bm{X}, \bm{X}))$.
	
	\item Prepare $W \in \mathbb{R}^{P \times L}$ for samples $\bm{Y}$: draw $P$ $L$-dim vectors from standard normal distribution, reject $\bm{0}$ vector, draw more samples if rejection happened, and normalize each vector.
	
	\item Generate noise-free $\bm{Y}$: $\bm{Y}_e = \bm{G}_e @ W^T$, where $e=1, ..., E$, $@$ is the matrix multiplication operator, and $W^T$ is the transpose of $W$.
	
	\item Add gaussian noises to $Y$ and $Z$ with specified noise levels $\sigma_p = 0.4$ and $\sigma_{safety}=0.05$.
\end{enumerate}

Now we have datasets $\mathcal{D}_e = (\bm{X}, \bm{Y}_e, \bm{Z}_e)$ for $e=1, ..., E$.
We can pick the first $N_{training}$ as training samples and the rest as test samples.
This is equivalent to random data split because (MO)GP models are permutation invariant (i.e. data-shuffle invariant, which makes random selection the same as shuffling $\bm{X}$ at step 1) and because $\bm{X}$ are drawn randomly without being sorted.

The experiment starts with $N_{sum}=40$ ($10$ for each output), and we repeat the experiment with 30 different seeds.
In this dataset, the RMSE and log likelihood are evaluated on noisy test data.
The noise-free data are not accessible throughout  the experiment.

For all of the 30 experiments, we compute the 20\%-quantile of $\bm{Z}_1 \cup ... \cup \bm{Z}_E$, denoted by $z_{0.2}$, and set the data safe when $p(h(\bm{x}) > z_{0.2}) > 0.95$. 

\paragraph{EngE dataset}\label{supp_section-exp_detail_OWEOdata} \ 

This dataset has 8 output channels including 2 temperature channels and 6 chemical substances emitted from a gasoline engine.
All of the data were measured from a warm engine and were split into training and test datasets.
The 2 temperature channels are highly correlated with Pearson correlation coefficient close to 0.98.
The datasets were normalized such that each input or output channel of the training set has mean 0 and variance 1 with negligible numerical error.
Therefore, it is suitable for a pool-based active learning algorithm.
In addition, the engine is a dynamic system, i.e. outputs depends on inputs of not only current time points but also past histories, and a sequence of data is used together in order to make accurate predictions.
In order to reflect the dynamic aspects, the dataset is available with a history considering nonlinear exogenous (NX) structure, concatenating the relevant past points into the inputs.
Inputs of this dataset have originally 5 channels (i.e. $\bm{x} \in \mathbb{R}^5$), and individual channels may have different history structures.
With the history concatenation, the inputs have 14 dimensions.

The data were measured with high sampling frequency.
The training set has in total around 247 thousand points, but in practice if we train a sparse MOGP model~\citep{vanderwilk2020framework}, the performances saturate with few thousand of randomly selected data (in this case we did not consider any safety constraint, which could deteriorate the performance).
Our safe AL experiment achieved a test RMSE of 0.85 with roughly 100 observations ($N_{sum}$) under safety consideration, while the saturation we achieved was 0.65, using much more observations and at least hundreds of inducing points leading to a larger memory requirement.

In the main experiment, we start from 48 data, 24 for HC, 24 for O2, and all 48 for the safety values.
The 3 frameworks start with the same initial data.
In each AL iteration, either HC or O2 is queried together with the corresponding temperature value.
We use seed 123 to randomly generate 30 sets of initial data, and perform 30 experiments with these initial sets, each with an individual seed (affect the random selection benchmark). Both the RMSE and test log likelihood show that our methods perform better than the competitors (figure~\ref{supp_fig-RMSE_log_dens}).

For the safety threshold $z_{max}$, the 80\%-quantile of this temperature channel in the processed training dataset is 1.0075.
We round this number to 1 as the threshold for the experiment.
Notice that, 20.55\% of the data is unsafe with $z_{max}=1$.
The safety constraint in this experiment is thus $p(h(\bm{x}) \leq 1.0) > 0.95$.

For some systems, it might be relatively easy and cheap to collect observations of all channels.
To investigate the performance of AL\_MOGP for this situation, we conduct the following ablation study.

\section{Ablation Study}\label{supp_section-ablation}

We perform the same experiments as described above on fully observed data.
In a partially observed output (POO) setup, we start from $N_{sum}$ input points, $N_{sum}/P$ output points for each output $\{y_{pn} | n=1, ..., N_{sum}/P\}$, $N_{sum}$ safety values $\{z\}$, and the safe AL proceed by querying $\{((\bm{x}_a, p), y_{pa}, z_a)\}$.
In a fully observed output (FOO) setup, we start from $N_{sum}/P$ input points, $N_{sum}/P$ $P$-dimensional output points $\{\bm{y}_{n} | n=1, ..., N_{sum}/P\}$, $N_{sum}/P$ safety values $\{z\}$, and the safe AL proceed by querying $\{(\bm{x}, \bm{y}, z)\}$ (i.e. in each AL iteration, $\bm{Y}$ gain $P$ points, notated in a POO manner, and $\bm{Z}$ gain 1 point).

We compare the RMSE of model $\bm{f}$ under POO and under FOO, given the same number of training points.
We perform the experiments on the simulation datasets, and figure~\ref{supp_fig-POOvsFOO_sims} shows that our AL\_MOGP under POO is the most data-efficient.
This is as expected because, in addition to MOGP's capability of correlation learning, POO provides more flexibility of exploration than FOO, given fixed number training points (or given fixed budget in a real application).
Here our datasets have similar level of uncertainty for different outputs by design.
When one of the outputs has much larger level of uncertainty than the others, our acquisition function for POO might tend to query mostly from this uncertain output, which we did not investigate in detail.

Notice that with fixed number of training outputs $\bm{Y}$, the number of observed safety values is less under FOO than under POO.
However, we do not compare the safety models under different setup, as our goal is to have a good safety control which is achieved by both POO (high model precisions shown in table~\ref{table-safety_toy}~\ref{table-safety_GP}~\ref{table-safety_OWEO}) and FOO (high precisions, not shown in this paper).
\cite{10.1007/978-3-319-23461-8_9} provides more insight in safety guarantee.

\end{document}